\documentclass[electronic]{vgtc}              

\graphicspath{{figures/}{pictures/}{images/}{./}} 

\usepackage{times}                     

\usepackage{tabu}                      
\usepackage{booktabs}                  
\usepackage{mwe}                       
\usepackage{listings}          

\usepackage{xcolor}
\usepackage{graphicx}
\usepackage{subcaption}
\usepackage{float}
\usepackage{comment}
\usepackage{wrapfig}

\usepackage{calc}

\newlength\myheight
\newlength\mydepth

\newcommand*\wordimg[1]{%
  \settototalheight\myheight{Xygp}%
  \settodepth\mydepth{Xygp}%
  \raisebox{-\mydepth}{\includegraphics[height=\myheight]{#1}}%
}

\newcommand{\osflink}{\href{https://osf.io/wsrzm/overview?view_only=64b8d81b21344085954f78192ae0d19d}{\texttt{osf\discretionary{}{.}{.}io\discretionary{/}{}{/}wsrzm}}\xspace}

\newcommand{\sm}{Supl. Mat.\xspace}

\usepackage{mathptmx}                  

\definecolor{jccolor}{RGB}{0, 0, 0}
\newcommand\jc[1]{{\color{jccolor}#1}}

\newcommand{\eg}{e.\,g.}
\newcommand{\ie}{i.\,e.}

\newif\ifverA
\verAtrue

\newcommand{\mytitle}{A Stevens's Power Law Check-up of GPT-5.5's \\ Implicit Reading  of Visual Encoding}

\onlineid{6113}

\vgtccategory{Research}

\vgtcinsertpkg

\title{\mytitle}

\author{Kaichun Yang\thanks{e-mail: yang.5853@osu.edu} %
\and Jian Chen\thanks{e-mail: chen.8028@osu.edu} %
}
\affiliation{\scriptsize The Ohio State University}

\abstract{We adapt Stevens's power law to measure the \textit{implicit} ability of AI models to read visualizations, which can reveal the built-in perceptual mechanisms of algorithmic models. In our pilot study, AIs see no legend. In the color conditions, no colormap name is provided either. GPT-5.5 first views a reference visual representation and estimates its magnitude, then estimates the magnitude of each subsequent image of the same representation relative to that reference. 
Our evaluation of twelve visual variables makes how algorithmic models read visual encodings measurable, comparable with human perception, and more transparent to humans. \osflink has source code, images, magnitude data samples, and analyses.}

\keywords{AI evaluation, behavior quantification, visual perception, Stevens's power law.}

\begin{document}



\firstsection{Introduction}

\maketitle

Quantifying artificial intelligence (AI) algorithms is exceptionally challenging with regard to their behaviors concerning data.
Multimodal large language models (MLLMs) are increasingly used to interpret charts~\cite{Zhang:2024:TinyChart} and solving problems with visual representations in robotics, natural scenes in physical AI. In these cases, a model may need to read values from, and compare or reason over graphical representations. However, real-world images of varying forms may not always present legend, or are lack of design consistency~\cite{chen2021vis30k}. 
For these reasons, we must make the algorithmic behaviors transparent, \ie, to discover how the machine represents visualizations. This is also aligned to problem solving, urged by Simon~\cite{simon1996sciences} requires knowing and defining the observers' representation of the domain-specific problems.
While prior graphical perception research has established how accurately AIs perform humans' tasks~\cite{ClevelandMcGill:1984:GraphicalPerception, Mackinlay:1986:ADG}, \eg,  ratio estimates~\cite{haehn2018evaluating}, more recent work has begun to evaluate AIs' mechanisms using rigorous sampling-based~\cite{Jiang:2026:SR} and visual literacy tests~\cite{hong2025llms}. However, these evaluation experiments typically use standard visualization images for human observers and implicitly assume that AIs perceive a visualization image as humans do. 
AI's accuracy in these cases may instead come from 
pattern matching, such as recalling familiar chart styles or reading verbal description or just fabricating a visual input that was never given in so-called mirage effects~\cite{schaeffer2023emergent, asadi2026mirage}.
Methods to quantifying AIs' implicit perceptual behaviors toward encoded data 
remains unattested.

\begin{figure}[t]
    \centering
    \includegraphics[
        width=\columnwidth,trim={4 2 3.5 3}, clip
    ]{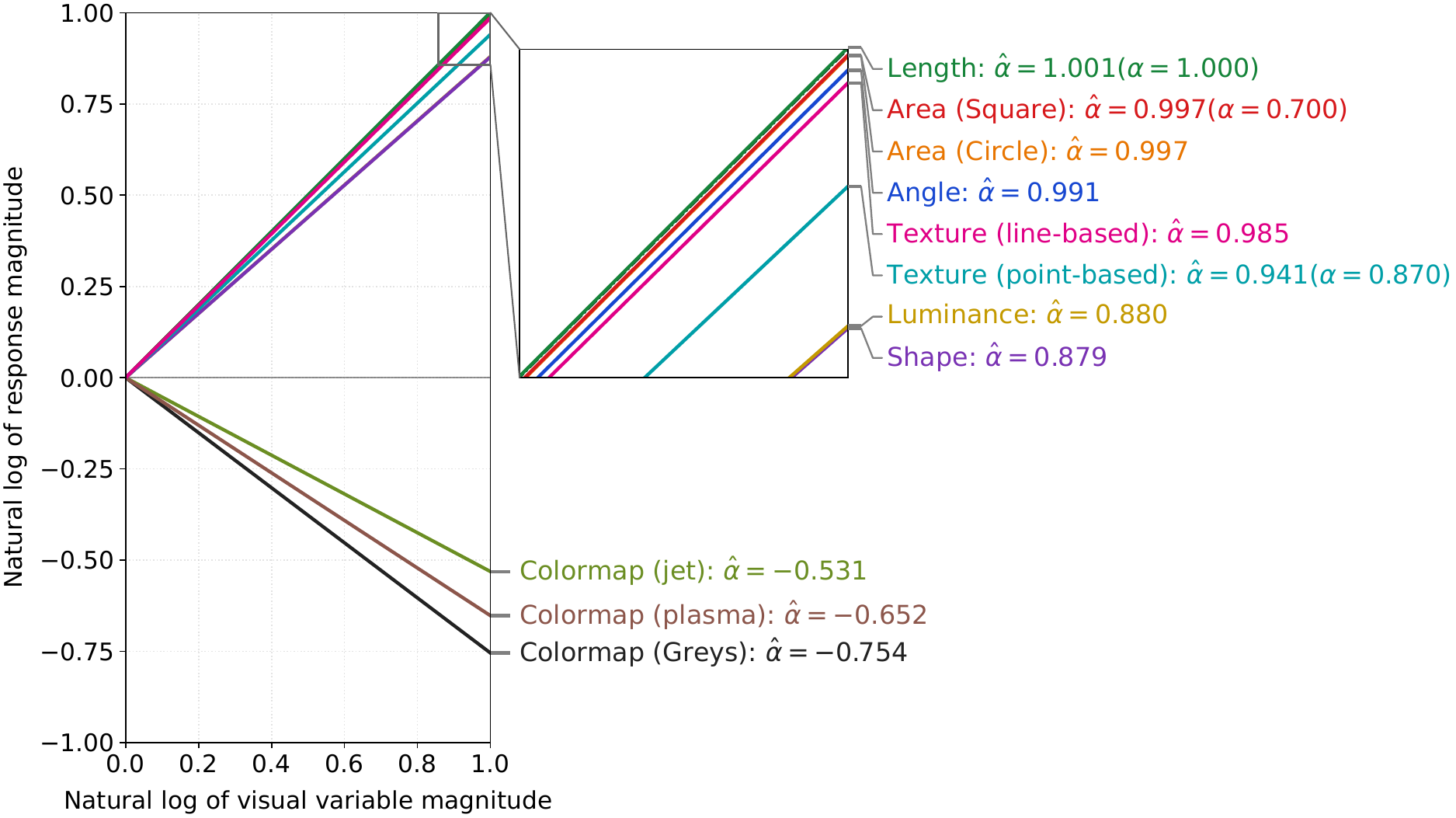}
    \caption{\textbf{A new psychophysical measurement of how neural network models respond to visual variables.}
We adapt Stevens's measurement of humans' subjective sense of physical stimulus intensity to measure how machines (GPT-5.5) respond to visual variables, \eg, length, area, color, and texture, and so on.  
The $\alpha$ value describes how observers perceive magnitude changes as the intensity of a visual variable changes. The model's response scaling is compressive when $\alpha<1$, linear when $\alpha=1$, and expansive when $\alpha>1$.
$\hat{\alpha}$ denotes the exponent for the model counterpart, and $\alpha$ for human observers where available.
\textbf{Observations.} 
GPT-5.5's responses scaled from linear to mildly compressive across visual variables. The three colormaps (\textit{Greys}, \textit{plasma}, and \textit{jet}), 
had negative $\alpha$, indicating that AIs reversed the direction of these maps. 
}
    \label{fig:power_law}
\end{figure}

\begin{figure*}[!t]
    \centering
    \setlength{\tabcolsep}{5pt}
    \setlength{\fboxsep}{0.5pt}
    \setlength{\fboxrule}{0.4pt}

    \captionsetup[subfigure]{
        font=scriptsize,
        labelfont=bf,
        labelsep=space,
        justification=centering,
        skip = 2pt
    }

    \begin{tabular}{@{}cccccc@{}}
        \begin{subfigure}[t]{0.16\textwidth}
            \centering
            \fbox{%
                \includegraphics[
                    trim={120 120 120 120},
                    clip,
                    width=0.75\textwidth
                ]{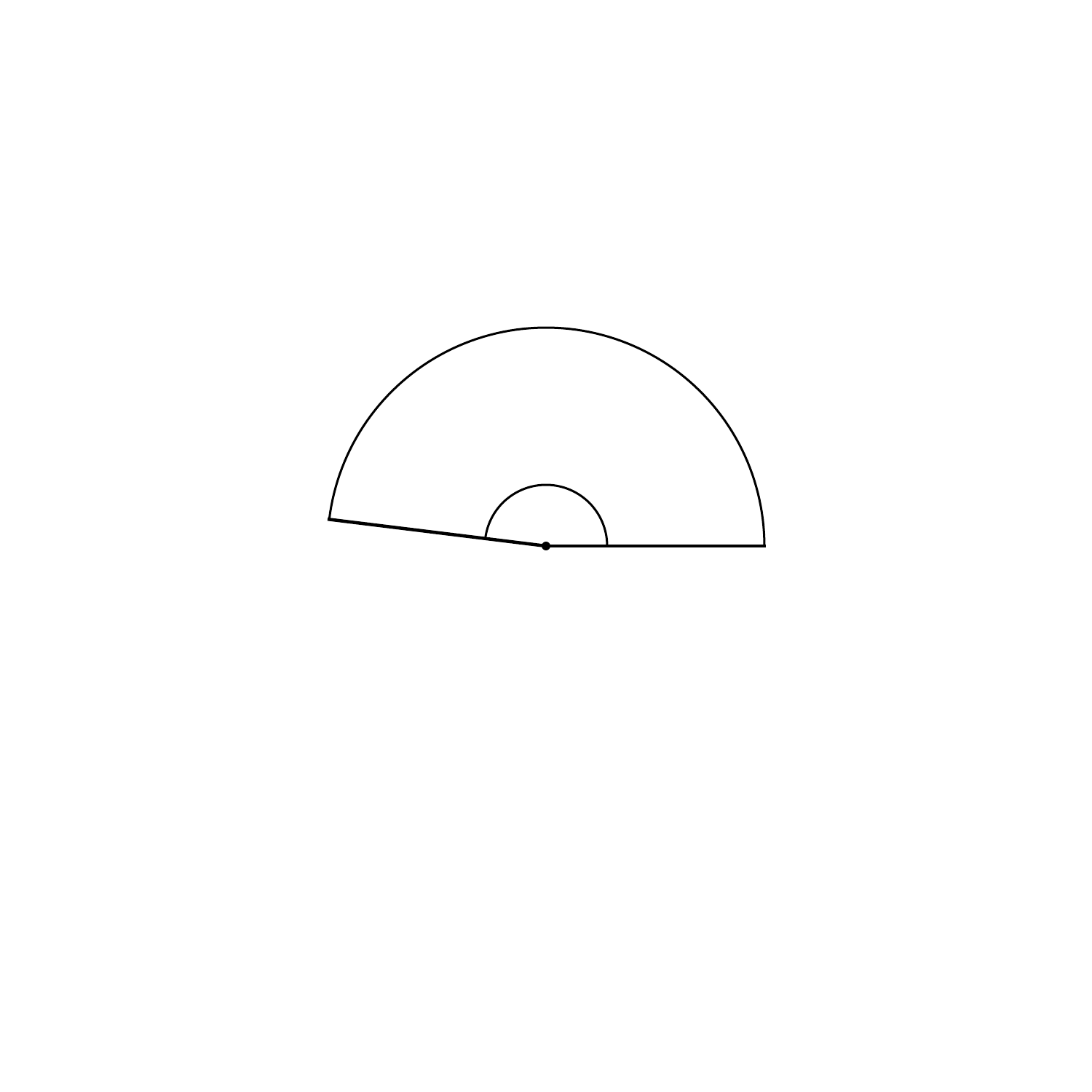}%
            }
            \caption{Angle}
            \label{fig:variable_angle}
        \end{subfigure}
        &
        \begin{subfigure}[t]{0.16\textwidth}
            \centering
            \fbox{%
                \includegraphics[
                    trim={120 120 120 120},
                    clip,
                    width=0.75\textwidth
                ]{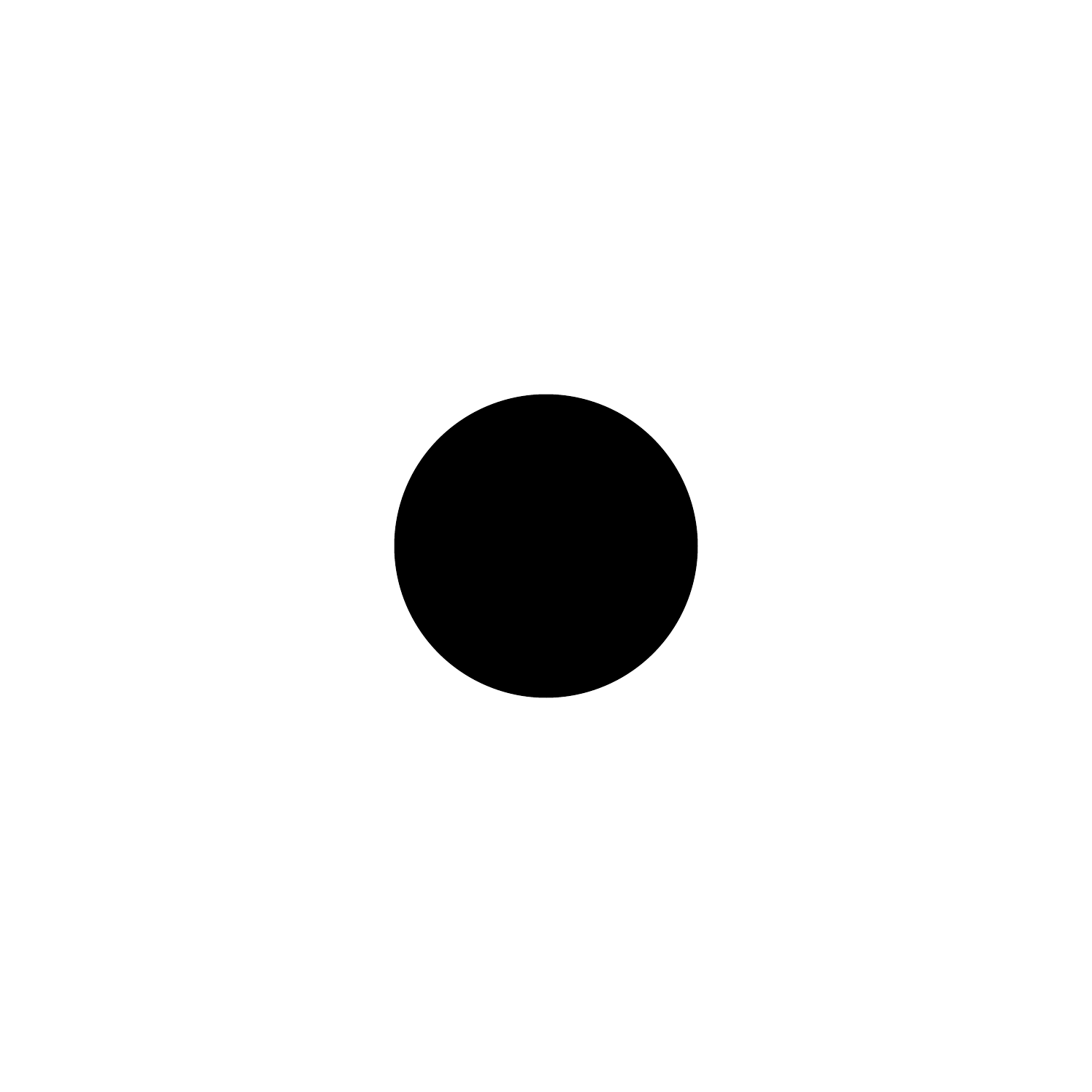}%
            }
            \caption{Area (Circle)}
            \label{fig:variable_area_circle}
        \end{subfigure}
        &
        \begin{subfigure}[t]{0.16\textwidth}
            \centering
            \fbox{%
                \includegraphics[
                    trim={120 120 120 120},
                    clip,
                    width=0.75\textwidth
                ]{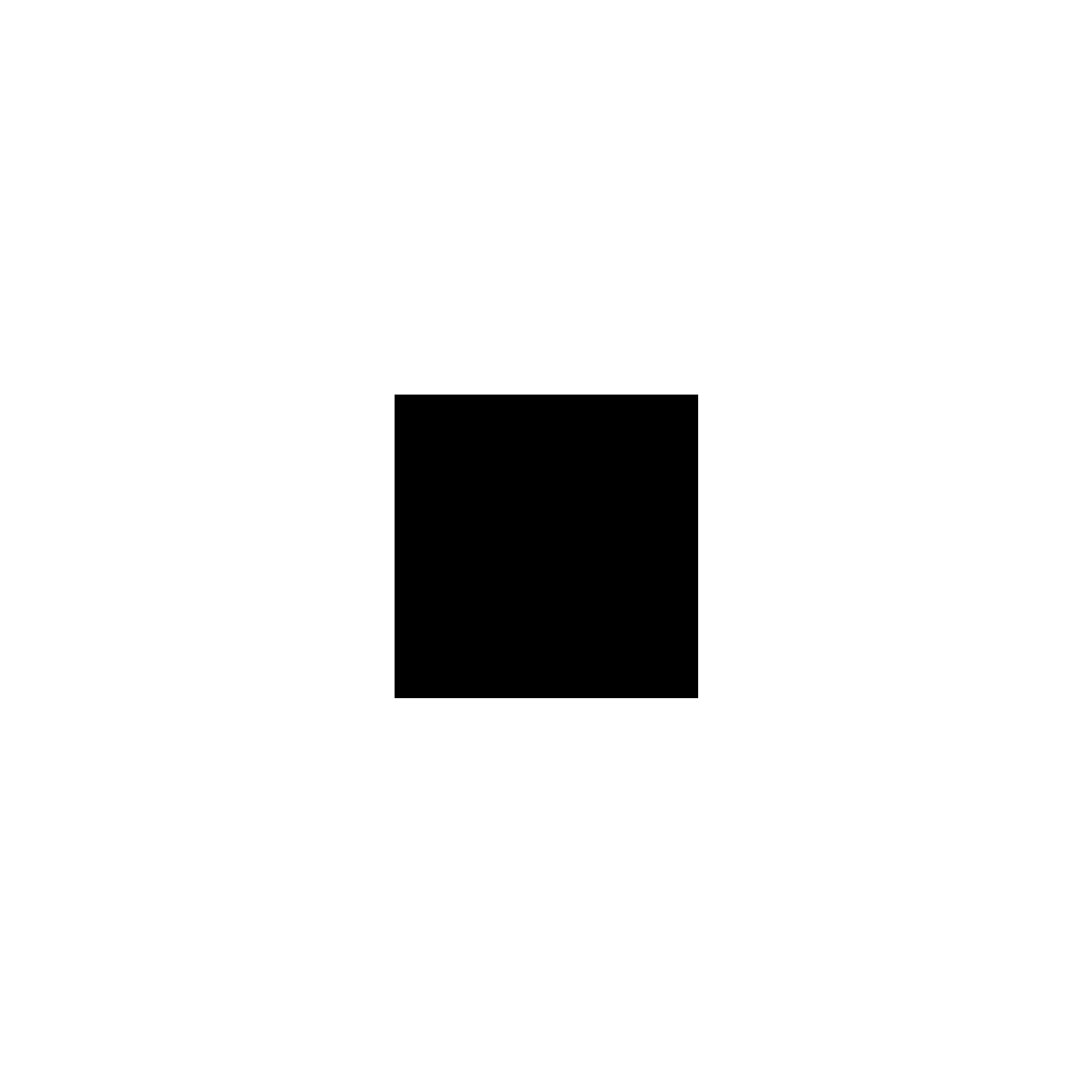}%
            }
            \caption{Area (Square)}
            \label{fig:variable_area_square}
        \end{subfigure}
        &
        \begin{subfigure}[t]{0.16\textwidth}
            \centering
            \fbox{%
                \includegraphics[
                    trim={120 120 120 120},
                    clip,
                    width=0.75\textwidth
                ]{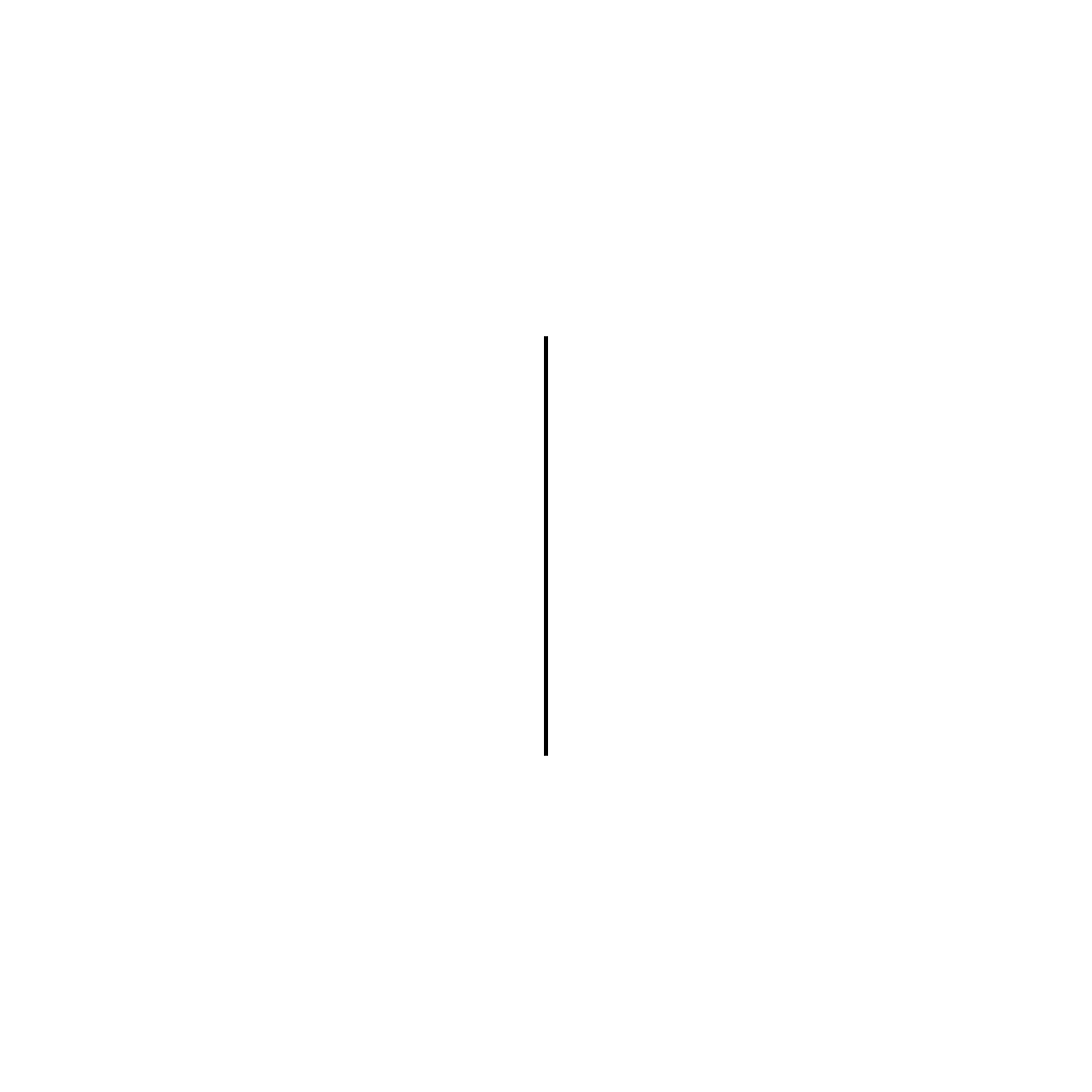}%
            }
            \caption{Length}
            \label{fig:variable_length}
        \end{subfigure}
        &
        \begin{subfigure}[t]{0.16\textwidth}
            \centering
            \fbox{%
                \includegraphics[
                    trim={120 120 120 120},
                    clip,
                    width=0.75\textwidth
                ]{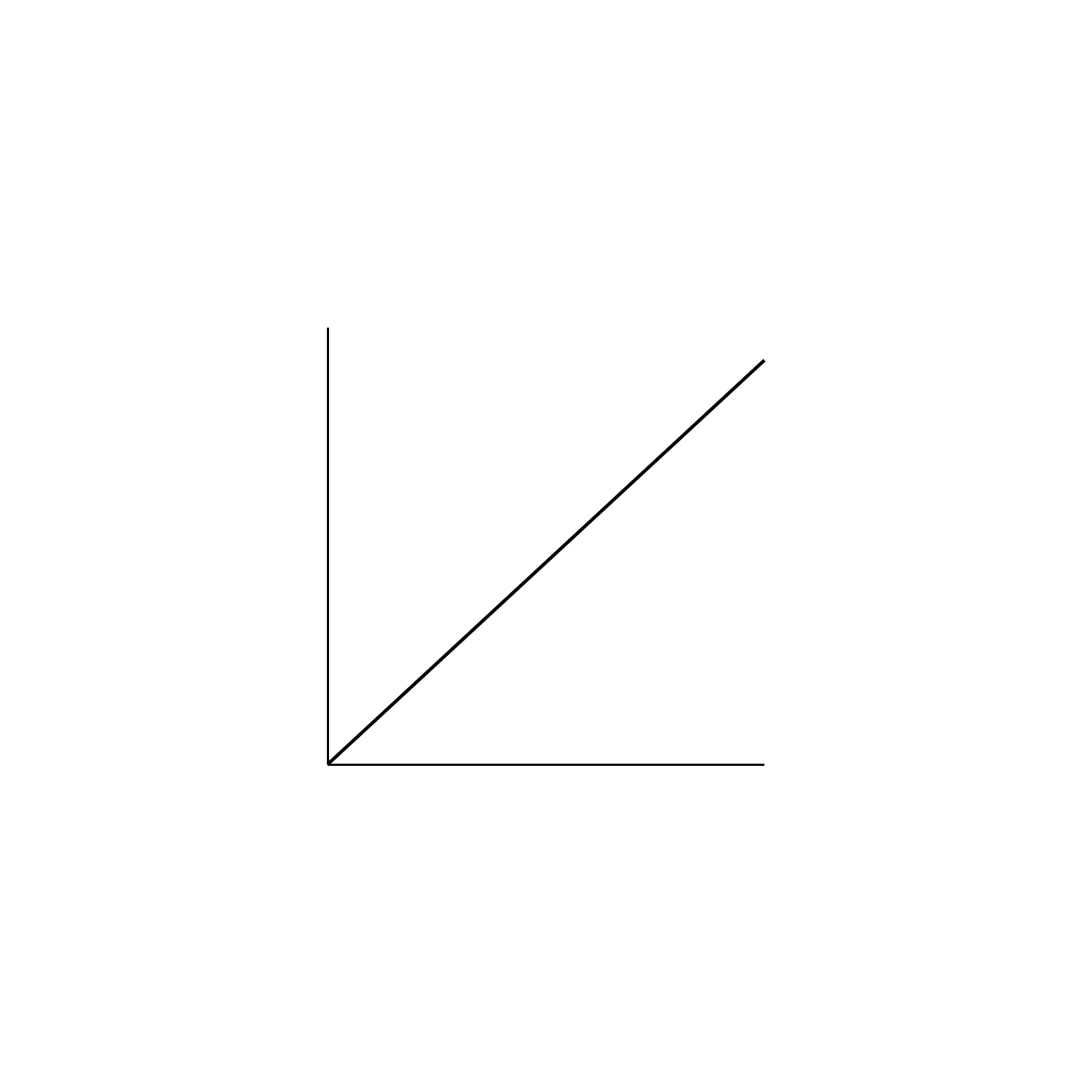}%
            }
            \caption{Slope}
            \label{fig:variable_slope}
        \end{subfigure}
        &
        \begin{subfigure}[t]{0.16\textwidth}
            \centering
            \fbox{%
                \includegraphics[
                    trim={120 120 120 120},
                    clip,
                    width=0.75\textwidth
                ]{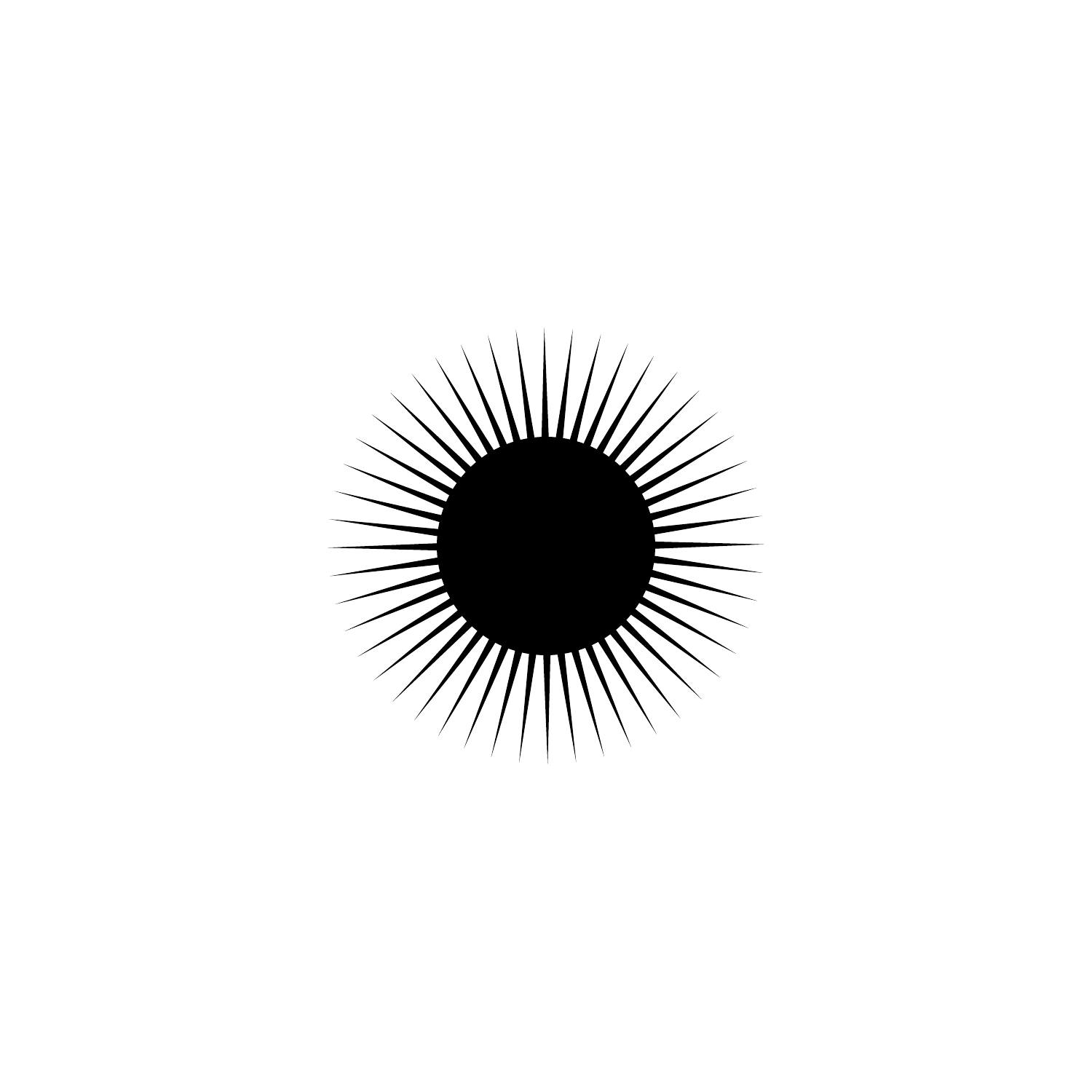}%
            }
            \caption{Shape}
            \label{fig:variable_shape}
        \end{subfigure}
        \\[20pt]

        \begin{subfigure}[t]{0.16\textwidth}
            \centering
            \fbox{%
                \includegraphics[
                    trim={120 120 120 120},
                    clip,
                    width=0.75\textwidth
                ]{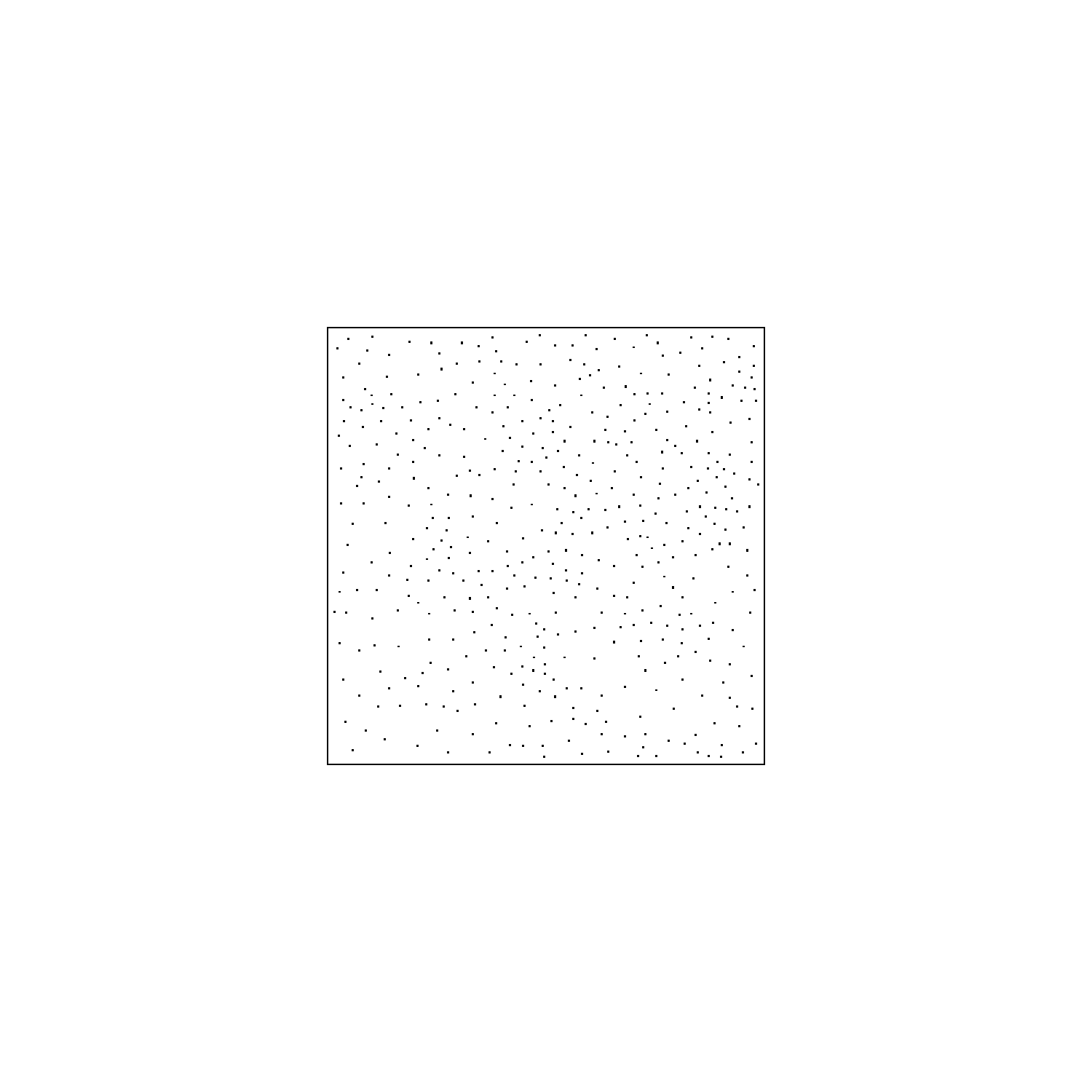}%
            }
            \caption{Texture (point-based)}
            \label{fig:variable_texture_dot}
        \end{subfigure}
        &
        \begin{subfigure}[t]{0.16\textwidth}
            \centering
            \fbox{%
                \includegraphics[
                    trim={120 120 120 120},
                    clip,
                    width=0.75\textwidth
                ]{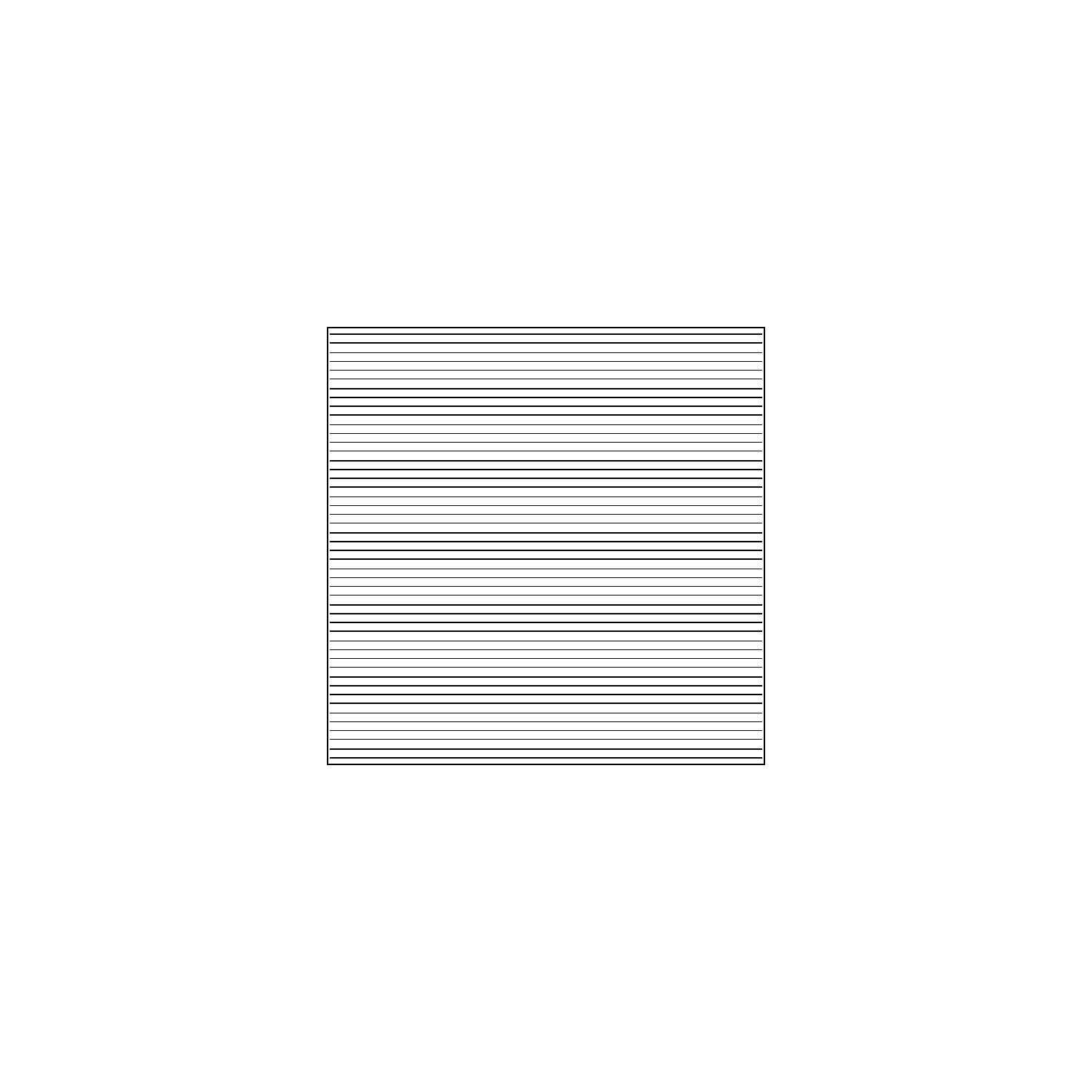}%
            }
            \caption{Texture (line-based)}
            \label{fig:variable_texture_line}
        \end{subfigure}
        &
        \begin{subfigure}[t]{0.16\textwidth}
            \centering
            \fbox{%
                \includegraphics[
                    trim={120 120 120 120},
                    clip,
                    width=0.75\textwidth
                ]{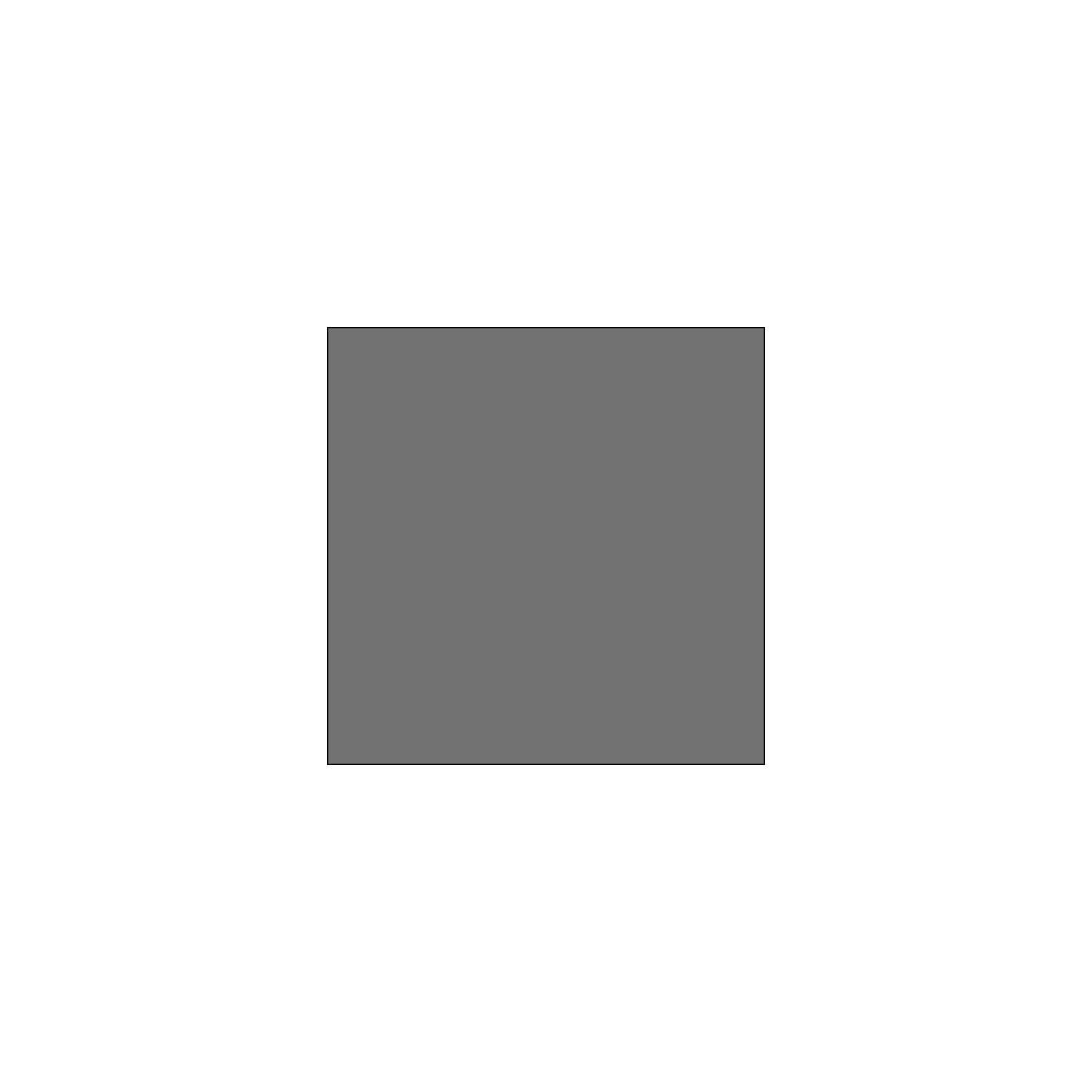}%
            }
            \caption{Luminance}
            \label{fig:variable_luminance}
        \end{subfigure}
        &
        \begin{subfigure}[t]{0.16\textwidth}
            \centering
            \fbox{%
                \includegraphics[
                    trim={120 120 120 120},
                    clip,
                    width=0.75\textwidth
                ]{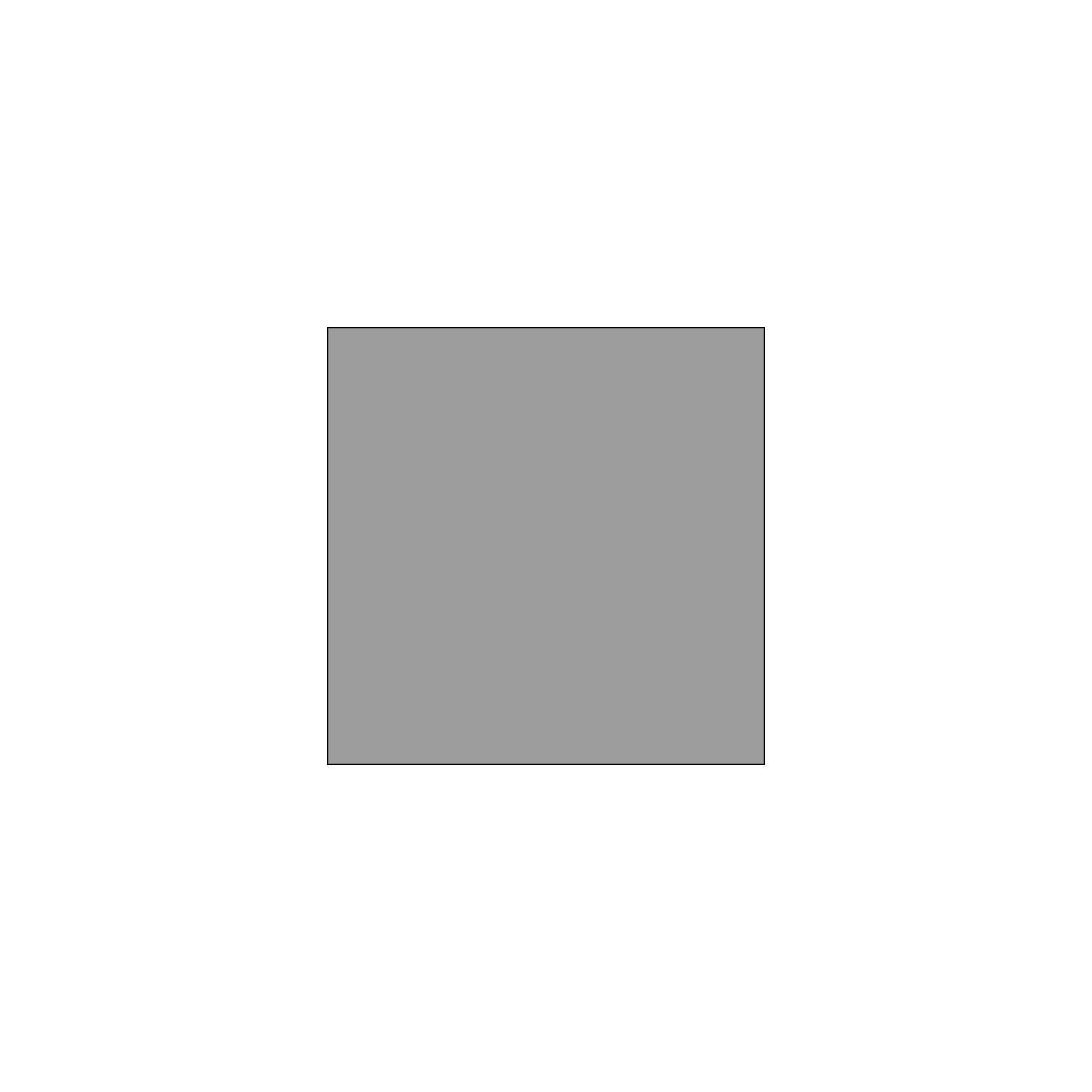}%
            }
            \caption{Colormap (Greys)}
            \label{fig:variable_colormap_Greys}
        \end{subfigure}
        &
        \begin{subfigure}[t]{0.16\textwidth}
            \centering
            \fbox{%
                \includegraphics[
                    trim={120 120 120 120},
                    clip,
                    width=0.75\textwidth
                ]{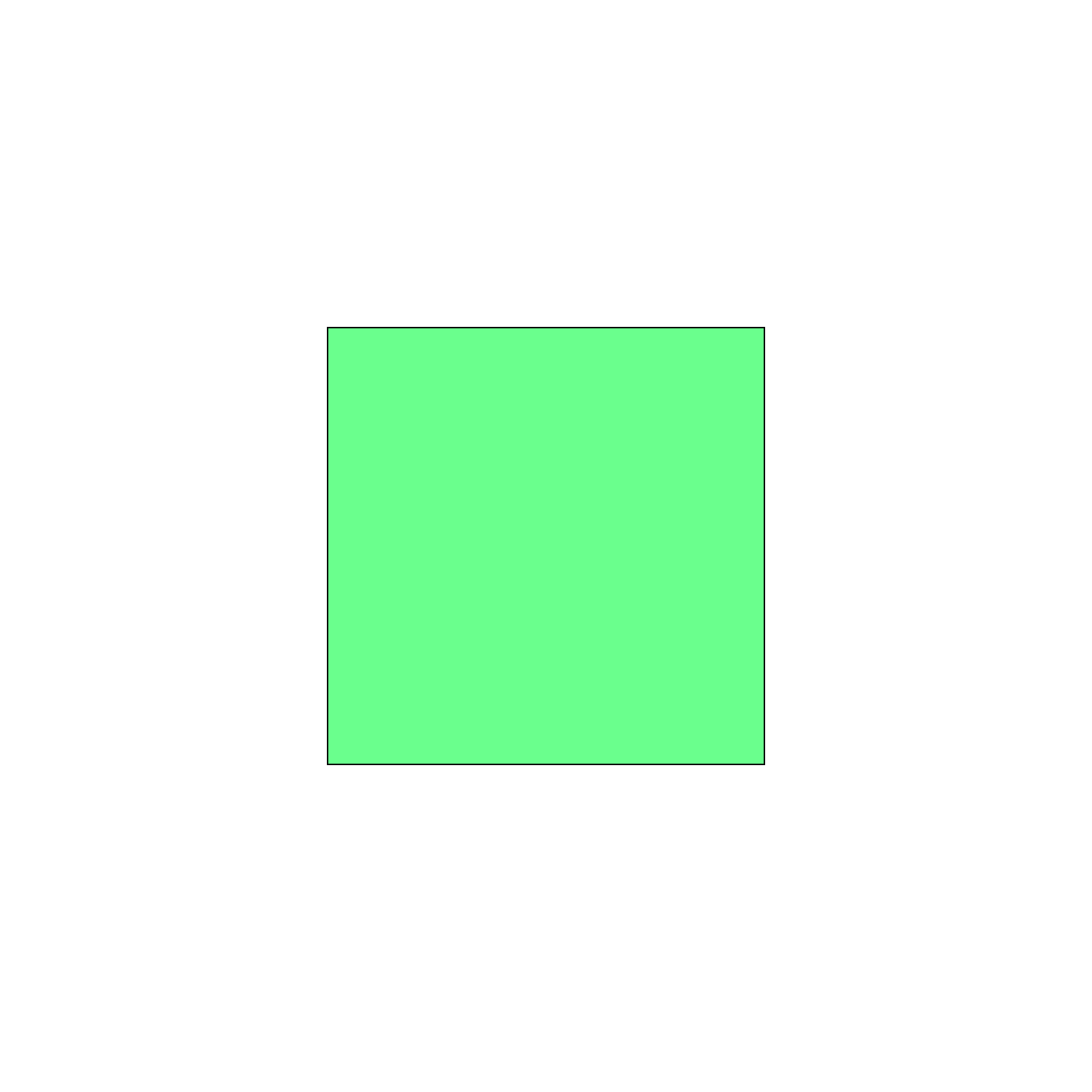}%
            }
            \caption{Colormap (jet)}
            \label{fig:variable_colormap_jet}
        \end{subfigure}
        &
        \begin{subfigure}[t]{0.16\textwidth}
            \centering
            \fbox{%
                \includegraphics[
                    trim={120 120 120 120},
                    clip,
                    width=0.75\textwidth
                ]{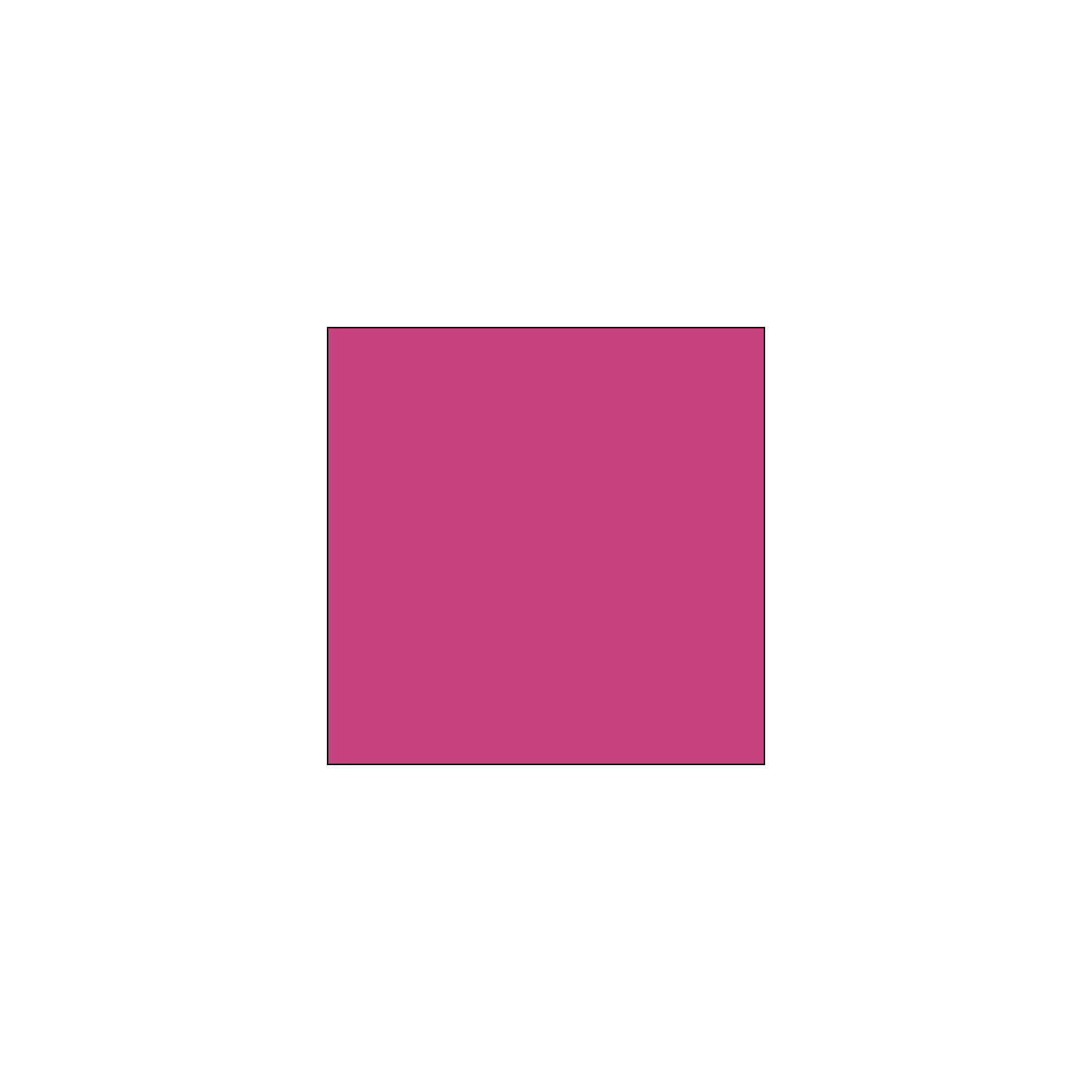}%
            }
            \caption{Colormap (plasma)}
            \label{fig:variable_colormap_plasma}
        \end{subfigure}
    \end{tabular}

    \caption{\textbf{Visual variable used in the experiment.}
    The twelve variants span angle, area, length, luminance, slope, shape, texture, and colormap-based visual representations. Each panel was generated using the input value \(v=48\) before the conversation. \(v \in [1, 100]\), each visual variable value linearly maps to value \(v\) except for Slope, such as Angle \(\in [4^\circ,\, 360^\circ]\), Area (Circle) \(\in [400\pi,\, 40\,000\pi]\mathrm{px}^2\), Area (Square) \(\in [1\,600,\, 160\,000]\mathrm{px}^2\), Length \(\in [8, 800]\mathrm{px}\), Shape \(\in [1, 100]\) spikes, Texture (point-based) \(\in [10, 1000]\) points, Texture (line-based) \(\in [1, 100]\) lines, Luminance \(L^* \in [1, 100]\). For each colormap, we divided the full colormap to 1000 equal interval, represented each interval by its midpoint color, and map \(v\) to every tenth interval.}

    \label{fig:stimulus-variants}
\end{figure*}

Our goal is to test the \textit{implicit} visual scaling mechanisms that AI algorithms use to read visual representations. We adapted Stevens's power law methods~\cite{Stevens:1957:PL}, which were originally developed to measure how humans perceived variable magnitude, to AIs.

\textbf{Method.} Unlike prior AI evaluations that replicate human visual stimuli, we asked AIs (here GPT-5.5) to estimate encoded visualization images directly (reference value) and then use this reference to estimate the target values relative to its own reference value, without a legend or any concepts.
Therefore, AIs could not perform pattern matching from our input images, and may only
 use familiar concepts they have been trained.
Because the same data quantity can be encoded into a set of visual variables, we measured twelve of these: angle, two areas, length, luminance, slope, shape, two textures, and three colormaps (\autoref{fig:stimulus-variants}) by asking a magnitude estimate from GPT-5.5. 
Our experiment used \textit{one-trial-per-run} protocol, each trial is presented in a new conversation, so earlier trials cannot influence later responses or all trials are independent.
We fit the responses with Stevens's power law
to examine how AIs read data in visualization images 
internally. 

\textbf{Results.}
We analyzed the pilot study results with two methods: Stevens's power law for modeling GPT-5.5's responses, and Cleveland and McGill's ranking of visual variables~\cite{ClevelandMcGill:1984:GraphicalPerception}.
Two findings emerged. 
Across twelve visual variables, GPT-5.5's magnitude responses scale approximately linearly with length, area, angle, and line texture, and are mildly compressive for point texture, luminance, and shape. For the three colormaps (\textit{jet}, \textit{plasma}, and \textit{Greys}), the responses have an inverse relationship with the data values mapped to three colormaps~(\autoref{fig:power_law}).
Also, its ranking of channels by errors differs from the human ranking.
This pilot study makes three main \textbf{ contributions}:
\begin{itemize}
\item \textbf{A psychophysical method for discovering the implicit scaling of GPT-5.5.} 
We measure how AIs perceive visual variables, using legend-free, reference-relative estimates that are less likely to reflect pattern matching against model's familiarity with charts.
\item \textbf{Twelve-channel reading results.} Across twelve visual variables, we show model's perceptual scaling and errors. 
\item \textbf{Findings for GPT-5.5.} These findings make its readings transparent to human observers. 
\end{itemize}

\section{Related Work}


\textbf{Understanding AI Model Behavior.}
Benchmark scores summarize AI performance over a set of tasks. However, an aggregate score does not always show how model responses vary across tasks or input conditions. 
Recent work argues that intelligence evaluation should measure behaviors and skill acquisition rather than only performance on a fixed task set~\cite{Chollet:2019:MI}. ARC-AGI-2 extends this direction through fine-grained evaluation signals and human-tested tasks~\cite{Chollet:2025:ARCAGI2}.

Jiang et al. proposed a data-domain sampling regime to assess CNN graphical perception in bar-chart ratio estimation~\cite{Jiang:2026:SR}. Their analysis examined sensitivity to differences between training and test distributions, stability under limited samples, and performance relative to human observers. In this work, instead of directly measuring behaviors, we model behaviors via psychophysics laws: 
we sample stimuli across controlled value ranges and analyze how the model's numerical responses change with the variable. This lets us treat MLLMs as AI observers and model behaviors by their response functions, relative error, and how both vary across visual variables and compare against human observers.
Together, these studies motivate evaluations that systematically vary inputs and observe responses rather than relying only on aggregated benchmarks.

\textbf{A Brief Review of Stevens's Power Law.}
Stevens invented a psychophysics method to test how observers scale input stimulus~\cite{Stevens:1956:DE,Stevens:1957:PL,StevensGalanter:1957:RS}. Standard stimulus has a ground truth reference value (for example, calling a tone of a given loudness ``100''), and an human observer assigned it a value (say calling the same loudness ``120''). Observers then assigned numbers to other stimulus intensities in proportion to their own standard. 
In this way, because observers respond across the whole stimulus range, the procedure measures the \textit{global} relationship between physical intensity and perceived magnitude. Stevens discovered this relationship as a power function, which is fit separately for each visual variable:
\[
\psi(I) = kI^{\alpha},
\]
where \(I\) is the ground-truth stimulus intensity, \(\psi\) is the observer (humans in Stevens and GPT-5.5 in ours)'s response, \(k\) is a proportionality constant, and \(\alpha\), is the power-law exponent, which is arguably the most important variable because it describes how the observer's response scales with stimulus intensity. 
Taking the logarithm of both sides gives
\[
\ln\!\bigl(\psi(I)\bigr) = \ln k + \alpha \ln I,
\]
so  estimating \(\alpha\) as the slope of a linear regression in log-log space indicates compressive reading scaling when \(\alpha<1\), linear reading scaling when \(\alpha=1\), and expansive reading scaling when \(\alpha>1\). Stevens measured many perceptual continua and found that exponents differ widely across them. For example, perceived line length is roughly linear which we also found from GPT-5.5's reading.

Weber's law has been used to measure just noticeable differences (JNDs), the smallest change in a stimulus that observers can reliably detect in visualizations~\cite{harrison2014ranking, Rensink:2021:VisualizationStimulus, RensinkBaldridge:2010:Correlation}.
That approach and ours are 
complementary, since 
Weber's law quantifies discrimination, whereas Stevens's law quantifies the form of the perceptual scale. We use Stevens's law because our question is not whether a model can distinguish two close values, but how its estimates of magnitude compress or expand across the 
range of each visual variable.



\section{Psychophysical Framework for Quantifying AI Observers}

We investigate 
whether
the relationship between 
data values encoded as visual stimuli 
and MLLM responses
follows Stevens's power law, and answer 
the following research question:

\begin{itemize}

\item\textbf{RQ.} How do the Stevens's power-law exponents fitted to MLLM responses differ across visual variables?


\end{itemize}

The independent variable is the visual variable and the dependent variable is the error.


\subsection{Task}
We use
magnitude estimation
task.
At the beginning of each conversation, the model was presented with an image using a visual variable and asked to assign it a numerical value (the reference), 
the model was subsequently presented with one target image of the same visual variable,
and asked to assign each target
a numerical value using the scale established by its response to the reference variable~(\autoref{fig:stimulus-variants}).
Each variable was shown in a separate image accompanied by a prompt. The prompts included instructions shared across all tasks together with information specific to the visual variable\ifverA~(\sm~\autoref{sec:prompts})\fi. 
No conversation history was shared between conversations.



\begin{figure}[!t]
    \centering
    \includegraphics[trim={0 50 0 0}, clip, width=\columnwidth]{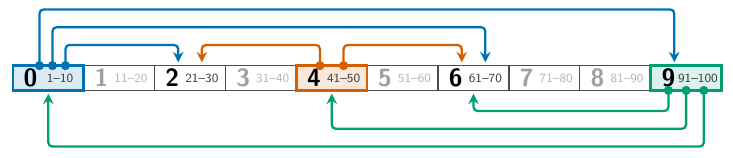}
    \caption{\textbf{
    Data sampling.} 
    Data values ranged from 1 to 100 and were divided into 10 bins. 
    Reference (R) and target (T) values were sampled from non-identical and non-adjacent bins. 
    This pilot study 
    used eight 
    reference-target (R-T) bin pairs, $R0\rightarrow\{T2,T6, T9\}$,
    $R4\rightarrow\{T2,T6\}$, and $R9\rightarrow\{T0,T4,T6\}$. Each R-T pair represents six different random sampled values.  
    }
    \label{fig:value-selection}
\end{figure}

\subsection{Study Design on Data Sampling}


Data 
values
ranged from \(1\) to \(100\), and  
were divided into ten non-overlapping bins: \([1,10]\), \([11,20]\), \(\ldots\), and \([91,100]\)~(\autoref{fig:value-selection}).
For this pilot study, we used 
a subset of
eight 
reference (R)-target (T) bin pairs: reference bin 0 (R0) paired with 
each of the target bins of 2 (T2), 6 (T6), and 9 (T9); R4 with T2 and T6; and R9 with T0, T4, and T6.
We excluded same-bin and adjacent-bin pairings to prevent the 
sampled R-T values from being too close, taken from Liu and Heer~\cite{liu2018somewhere}.
To avoid extremely small rendered stimuli that lead to large model prediction errors~\cite{Jiang:2026:SR}, no stimuli smaller than 32 pixels were used.  
For each of these eight R-T bin pairs, we randomly sampled six reference-target value pairs from the corresponding bins, resulted in
8$\times$6 = 48 
trials per visual variable. 

\subsection{Visualization Images}
The independent variable, visual variable, has 12 levels across eight different types: angle, area, color, length, luminance, slope, shape, and texture~(\autoref{fig:stimulus-variants}). 
All variable images used a \(1,000 \times 1,000\)-pixel square canvas with a white background.
Each image rendered one visual variable positioned at its geometric center.

\texttt{Angle}~(\autoref{fig:variable_angle})
was represented using two radial lines and inner and outer arcs. One radial line pointed horizontally to the right, while the other was rotated counterclockwise. A sampled 
data
value \(v\) was converted to degrees using \(a=\lceil3.6v\rceil\), producing angles $\in 
[15^\circ, 360^\circ]$. 
Data values with v$ < $4 were not used in the experiment to avoid 
inference errors introduced by small values~\cite{Jiang:2026:SR}.  This is consistently applied to all variables.
    
\texttt{Area}~(\autoref{fig:variable_area_circle} and~\ref{fig:variable_area_square})
was represented using solid black circles and squares. 
The area was 
mapped linearly to \( 400\pi v \,\mathrm{px}^2\) for circle area, and \( 1600 v \,\mathrm{px}^2\) for square area.

\texttt{Length}~(\autoref{fig:variable_length})
was represented using black vertical line segments with a fixed width of \(4\) px. 
For a 
data value \(v\), the rendered pixel length 
is \(l=8v\) px, or
pixel length 
range  $\in [32, 800]$.

\texttt{Slope}~(\autoref{fig:variable_slope})
was represented by a black line in a \(400\times400\) px L-shaped frame. \(v\) was mapped linearly from \(1\) to \(100\) to
an angle \(\theta\) $\in$ \(1.003^\circ\) to \(88.997^\circ\), and the slope was calculated as \(m=\tan(\theta)\). 

\texttt{Shape}~(\autoref{fig:variable_shape})
\jc{was treated as glyphs, inspired by star glyphs and glyph design principles for separability~\cite{borgo2013glyph}.}
Data were represented using solid black forms with evenly spaced spikes and a fixed base-circle radius of \(100\) pixels. Each spike was an isosceles triangle with a fixed height of
\(100\) px measured from its base chord, with circular arcs
retained between adjacent spikes. A sampled value \(v\) determined the number of spikes. Each shape was assigned a random rotation angle.

\texttt{Texture}~(\autoref{fig:variable_texture_dot} and~\ref{fig:variable_texture_line})
had two types, represented using line and dot textures inspired by He et al.~\cite{he2023design}, both drawn within a \(400\) px square region. For the line texture, 
a 
data 
value \(v\) corresponded to 
\(v\) approximately equally spaced horizontal lines with a width of \(1\) px. For the dot texture, \(v\) corresponded to \(10v\) black dots of size \(2\times2\) px, whose positions were generated using random initialization and Lloyd relaxation~\cite{Lloyd:1982:LSQ}, which reduced clustering and spread the dots more evenly across the square.

\texttt{Luminance}~(\autoref{fig:variable_luminance})
was represented using fixed-size squares. Both the reference and target were \(400\) px squares and differed only in their fill lightness. 
A value \(v\) 
was mapped to L* in the CIELAB color space (L$\in$[0,100], fixed $a^*=b^*=0$) 
, and then converted 
the resulting $L^*a^*b^*$ color 
to RGB for rendering. 



\texttt{Color}~(\autoref{fig:variable_colormap_Greys},~\ref{fig:variable_colormap_jet}, and~\ref{fig:variable_colormap_plasma})
was represented using fixed-size \(400\times400\) px squares filled according to Matplotlib's \texttt{Greys}, \texttt{plasma}, and \texttt{jet} colormaps.
For a 
data value \(v\), we 
converted it to a colormap index \(k=10v\) and then extracted the corresponding color from a colormap containing \(1,000\) positions.

\section{Results}

We describe the power-law models and 
error analyses
across visual variables. 
The main take-home results are: (1)
across twelve visual variables, GPT-5.5's magnitude responses scale approximately linearly with length, area, angle, and line texture, and are mildly compressive for point texture, luminance, and shape. (2) For the three colormaps (\textit{jet}, \textit{plasma}, and \textit{grays}), the responses have an inverse relationship with the data values mapped to color. 

\subsection{Analysis Method}

We collected 
$48\times12$= 576 target responses across 
48 trials of 12 visual variables.

\textbf{Stevens's power-law analysis.}
In our notation, \(v\) denotes the stimulus magnitude \(I\) for a given visual variable, and \(\hat{v}\) denotes the numerical responses \(\psi(I)\) of GPT-5.5.
For each reference-target pair, we calculated the visual variable ratio 
\(S=\frac{v_t}{v_r}\)
and the corresponding model-response ratio
\(R=\frac{\hat{v}_t}{\hat{v}_r}\),
where \(v_t\) and \(v_r\) are the target and reference values of the corresponding visual variable, and \(\hat{v}_t\) and \(\hat{v}_r\) are the values assigned by GPT-5.5. According to Stevens's power law, the response to a visual variable value \(v\) can be expressed as \(\hat v = kv^\alpha\). 
If the reference and target responses share the same scaling factor, taking their ratios should give \(R = S^{\alpha}\), and \(\ln R =\alpha \ln S\).
Then we fitted Stevens's power law by applying an ordinary least-squares regression to
\(\widehat{\ln R}=\hat{\beta}_0+\hat{\alpha}\ln S\)
separately for each visual variable, where \(\hat{\beta}_0\) and \(\hat{\alpha}\) are the estimated intercept and exponent.
We denote the fitted exponent as \(\hat\alpha\).
Only finite and positive
ratios were included because the logarithm is undefined for zero ratios.

For angle, we replaced a reported value \(r\) with its circular complement \(360^\circ-r\) when the complement was closer to the corresponding ground-truth angle. Slope was excluded from the power-law analysis and was included only in the error analysis after its stimulus values were converted to the rendered 
slopes.

\textbf{Log absolute error analysis.}
Following Cleveland and McGill~\cite{ClevelandMcGill:1984:GraphicalPerception}, we calculated the log absolute error for each target:
\[
e= \log_2\left( \left| 100\frac{\hat{v}_t}{\hat{v}_r} -
100\frac{v_t}{v_r} \right| +\frac{1}{8} \right).
\]

We then 
used 
 Analysis of Variance (ANOVA) to test the main effect of visual variable on mean log absolute error. 
When we observe a significant difference, we 
used Tukey's Honest Significant Difference (HSD) post-hoc test to identify which
visual variables differed. 
Finally, we 
generated 1,000 bootstrap samples by resampling the valid trials with replacement with each bootstrap sample contains the same number of trials as the original sample. The $2.5^{th}$ and $97.5^{th}$ percentiles of the bootstrap distribution were used as the 95\% confidence interval of mean log absolute error.


\subsection{GPT-5.5 Stevens's Power-Law Fits}

\autoref{fig:power_law} shows the fitted power-law exponents for GPT-5.5. The exponents for length, the two area variables, angle, and line-based texture ranged from \(0.985\) to \(1.001\), indicating that their assigned values increased approximately linearly with stimulus intensity. Point-based texture showed 
response compression (\(\hat{\alpha}=0.941\)), while luminance and shape had the smallest positive exponents (\(\hat{\alpha}\approx0.880\)).
Three colormaps had negative exponents, ranging from \(-0.754\) for \textit{Greys} to \(-0.531\) for \textit{jet}. Thus, GPT-5.5 tended to assign smaller values as the numerical stimulus value increased.


\begin{figure}[!tp]
    \centering
    \includegraphics[width=\columnwidth,trim={90 7 10 40},
    clip]{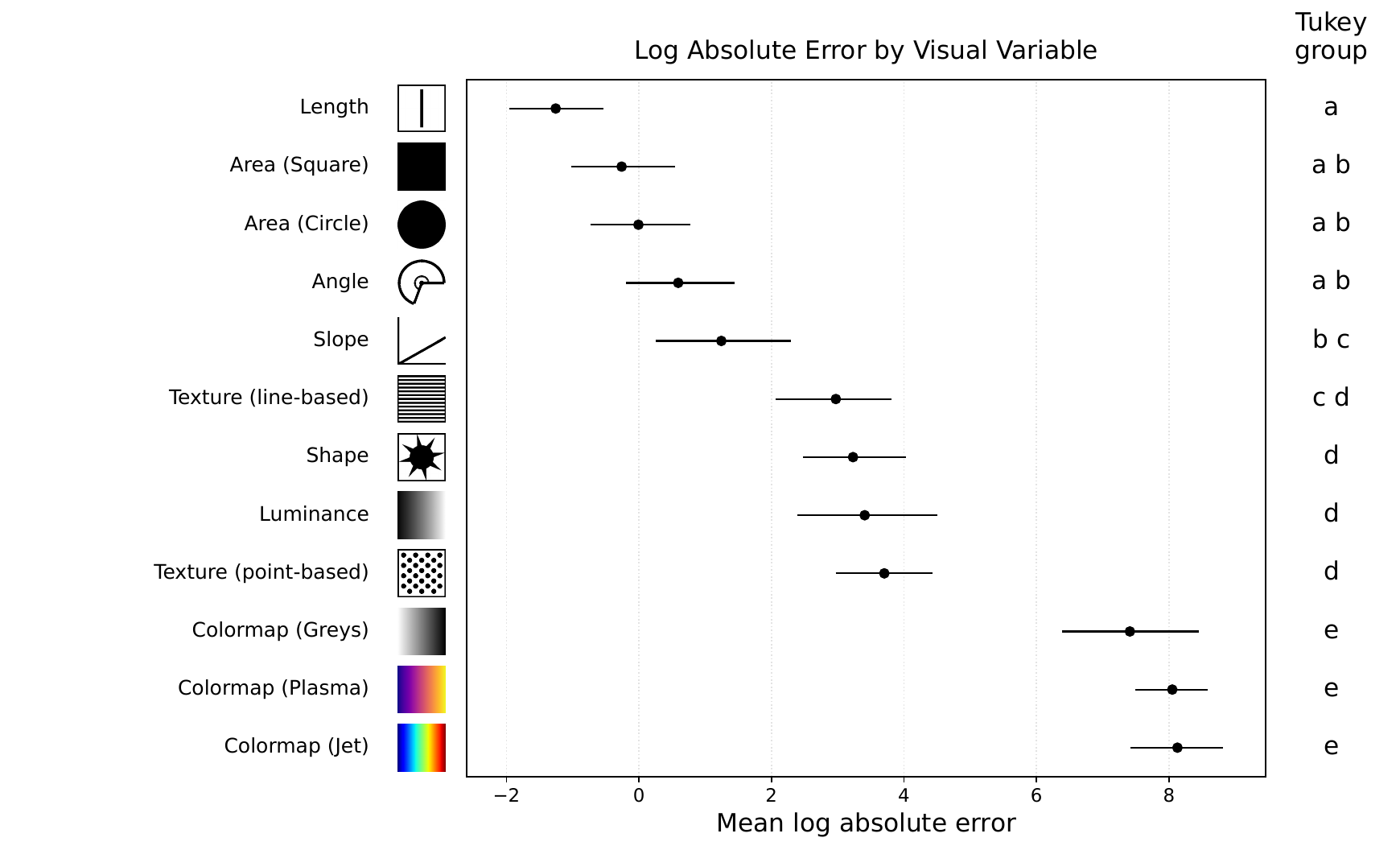}
    \caption{
    \textbf{Measurement Method II: Errors across visual variables.} Points show the mean log absolute error and bars show 95\% bootstrap confidence intervals. Letters indicate Tukey HSD groups; variables sharing a letter do not differ significantly in error.}
\label{fig:relative_err_attr}
\vspace{-2.5ex}
\end{figure}

\subsection{Ranking Visual Variables by Error}

In this analysis, we exclude 5 trials of colormap \textit{Greys}, 15 trials of colormap \textit{jet} and 1 trial of shape visual variables which GPT-5.5 assigned zero to their reference value. 
We find visual variable
was
a significant main effect on log absolute error,
\(F(11,543)=55.136\) and \(p<0.001\). Length produced the lowest mean log absolute error, followed by the two area variables, angle, and slope. The three colormaps produced the largest errors and were assigned to the same Tukey group. Each colormap had a significantly higher error than every non-colormap variable~(\autoref{fig:relative_err_attr}).

\subsection{GPT-5.5's implicit numerical mapping of colormaps}
\begin{figure}[!tp]
    \centering
    \includegraphics[width=\columnwidth]{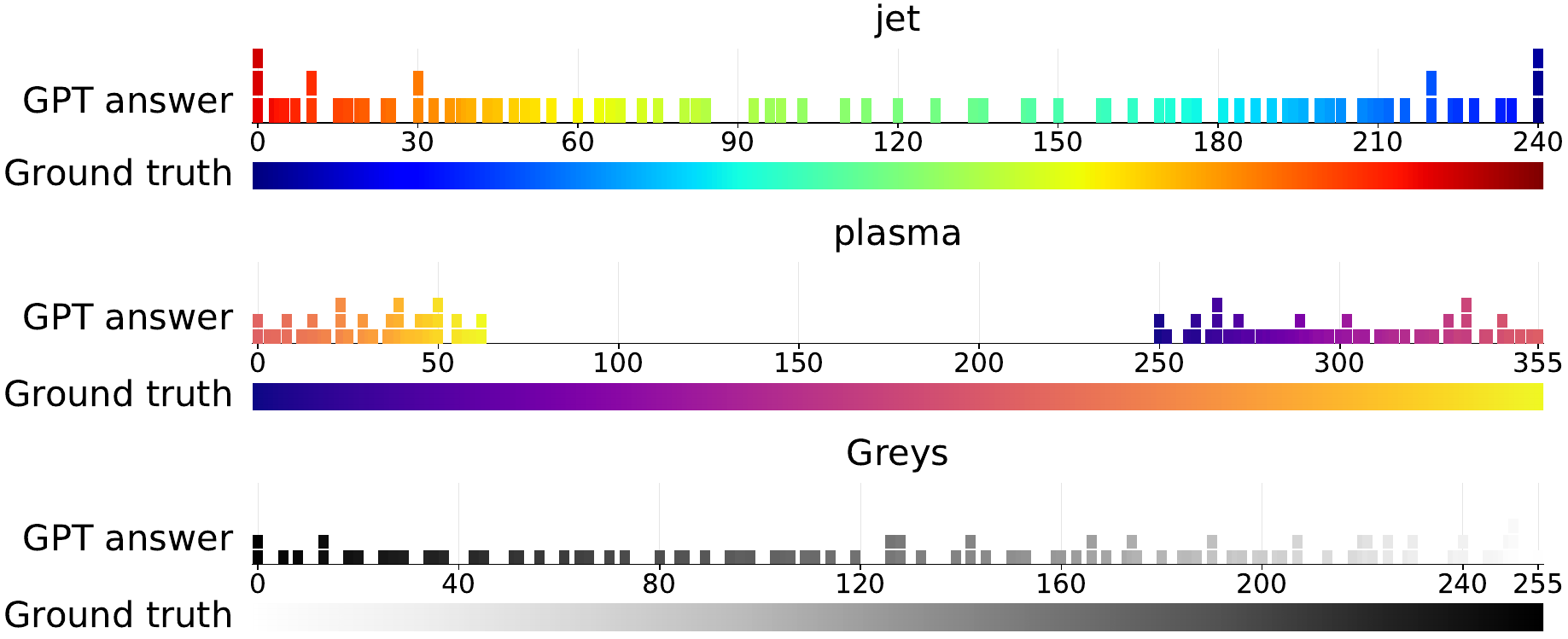}
    \caption{
    \textbf{Implicit colormap produced by the MLLM vs. the ground-truth colormap used in our experiment.} \textbf{Observations.} The MLLM wrapped color positions along the map, which caused the negative $\alpha$ values in the Stevens's power-law modeling. The three colormap names were not sent to the AIs in the prompt.
    \textbf{Observations.} GPT-5.5 implicitly mapped the provided single color patches to colormaps.}
    \label{fig:MLLM_jet_order}
    \vspace{-2.5ex}
\end{figure}

We suspected that GPT-5.5 using a colormap direction opposite to the numerical direction used to produce the input images since the $\alpha$ values were negative for the last three color maps. 
To further investigate this behavior, we conducted a follow-up full-sampling experiment to examine GPT-5.5's implicit numerical mapping of the colormaps. 
For each colormap, we selected the 50th sampled color as the reference stimulus and asked GPT-5.5 to assign it a number to color value. We then presented all remaining sampled colors as targets within the same conversation and asked the model to estimate their values relative to the reference. \autoref{fig:MLLM_jet_order} shows the resulting numerical mappings across the full colormap.
Without giving a legend and colormap name, GPT-5.5 read colors as hue angle (jet: 0–240; plasma: two clusters split by the 360° wrap-around) or gray level (Greys: 0–255) rather than colormap position. This reverses the stimulus direction for jet and Greys; for plasma it yields a negative fit despite preserved hue order. 

\section{Discussion}



\textbf{Measuring MLLMs as AI observers.} Using psychophysics methods gives us an
operational way to measure AIs' implicit mappings. Here, visualization serves as a stimulus domain for AI measurement science: the response
function and built-in scaling representations can be discovered using our method \textit{without a predefined mapping, a legend, or any stimulus-related information other than a single stimulus in the image, to test AI implicit representations.}

\textbf{The implicit wrapped color positions and lighter-as-higher bias}.
The most interesting finding is that, even without a colormap name, a legend, or any other color-related information beyond a single color patch, GPT-5.5 mapped the colors to numbers in a way that reflected the wrapped colors’ positions in the colormap, as it appears to read hue angle (jet and plasma) or gray level (Greys), even though some of these results were reversed.
Relative to humans, GPT-5.5 also mapped lighter-as-higher in Greys. This suggests that GPT-5.5 applies its own implicit color-to-number mapping. Designers may always need to provide a legend to obtain correct responses from GPT-5.5.
One possible reason is that lighter colors have higher luminance values in color space, for example white has the highest luminance.

\textbf{Limitations.}
The current analysis focused on one model, GPT-5.5. 
The pilot experiment also included a small  subset of the possible reference-target combinations, leaving fewer observations for modeling AIs. 
Future work should test additional MLLMs, more complete variable combinations, and different prompt conditions (\eg, real-world concepts embedded in colors). Comparing stimuli with and without legends may also show how much the AIs relies on visual information, explicit labels, or  semantic knowledge. 

\section{Conclusion}

Our new evaluation method reveals fundamental insights into the next steps of
interpreting MLLMs:
If a model can report the magnitude of an encoding
named only in text, with no legend in the image to match against, then it
carries a built-in sensory system that we can measure directly rather than infer
from task performance.
Our extension of Stevens's power law to model measurements shows that the implicit colormaps shows opposite to the one used to generate the visualization images. This result also support that having language alone might not be enough and visual information is necessary.
We therefore see that psychophysics experiment can be transparent about machine's problem solving gaps and emphasizes importance of having explicit context display such as legend may  be just as important for proper machine perception.





\section*{AI Use Statement}

\noindent 
GPT-5.6 Sol was used to assist in 
writing 
and debugging the plotting code for Figures 1, 4, and 5. It was not used to generate or modify the underlying data, analyses, or research findings. GPT-5.6 Sol and GPT-6 Astra helped debug and write experiment code and debug figure style code. All AI generated content was reviewed by the authors for accuracy.

\section*{Acknowledgments}
 
\noindent 
We thank reviewers and colleagues at the Visual Attention Lab at Harvard, led by Professor Jeremy M. Wolfe for their comments on this work. The work is supported in part by NIH U54CA287392, NSF IIS-1302755, NSF CNS-1531491, and NIST-70NANB13H181.
Any findings, conclusions, or recommendations in this work are solely the responsibility of the authors and do not necessarily represent the  views of the NSF, NIH, or the FDA.

\bibliographystyle{abbrv-doi}

\bibliography{template}

\newpage

\clearpage

\noindent\begin{minipage}{\columnwidth}
\makeatletter
\centering
\rmfamily\LARGE
\mytitle\\[1em]
\large{\textsc{Supplemental Materials}}
\makeatother
\end{minipage}
\vspace{0.4cm}


\appendix                                 

\lstdefinestyle{prompt}{
  basicstyle=\ttfamily\footnotesize,
  breaklines=true,
  breakatwhitespace=true,
  columns=fullflexible,
  keepspaces=true,
  showstringspaces=false,
  escapeinside={(*@}{@*)},       
}

\noindent{This supplemental material section provides our exhaustive description of the process for reproducible science.}

\section{GPT-5.5 configuration} 
\label{app:gpt5.5-config}

\begin{table}[htbp]
    \centering
    \small
    \caption{GPT-5.5 configuration used in the pilot study.}
    \label{tab:gpt5.5-config}
    \begin{tabular}{@{}ll@{}}
        \toprule
        Parameter & Value \\
        \midrule
        \texttt{model} & \texttt{gpt-5.5-2026-04-23} \\
        \texttt{reasoning.effort} & \texttt{low} \\
        \texttt{text.verbosity} & \texttt{low} \\
        \texttt{detail} & \texttt{original} \\
        \bottomrule
    \end{tabular}
\end{table}

\section{Prompts Uses} 
\label{sec:prompts}

\newenvironment{promptbox}{%
  \par\begingroup
  \ttfamily\footnotesize\raggedright
  \setlength{\parindent}{0pt}\setlength{\parskip}{1pt}%
}{\par\endgroup}

\newcommand{\promptlabel}[1]{\par\medskip\noindent{\bfseries\sffamily\small #1}\par\nopagebreak}

\newcommand{\promptTail}{%
  Briefly explain your reasoning before giving the answer.\par
  Then give your final numerical answer on the last line in exactly this format:\par
  FINAL\_ANSWER: \textless{}number\textgreater{}%
}


\newcommand{\promptentry}[5]{%
  \par\bigskip\noindent
  \textbf{\wordimg{#2}\ #3}\par\nopagebreak\smallskip
  \begin{wrapfigure}{r}{0.16\textwidth}
    \vspace{-1.5ex}
    \centering
    \fbox{\includegraphics[trim={120 120 120 120},clip,width=0.14\textwidth]{#1}}
    \vspace{-2ex}
  \end{wrapfigure}
  \promptlabel{Reference prompt}
  \begin{promptbox}
    #4\par
    \promptTail
  \end{promptbox}
  \promptlabel{Target prompt}
  \begin{promptbox}
    #5\par
    \promptTail
  \end{promptbox}
  \par\medskip
}

\section*{Example stimuli and prompts}
\label{fig:input-overview}

Each entry shows the visual variable image (right) with the reference and target prompts used for that variable.

\promptentry
  {all_figures/stimuli/angle_id_48.pdf}
  {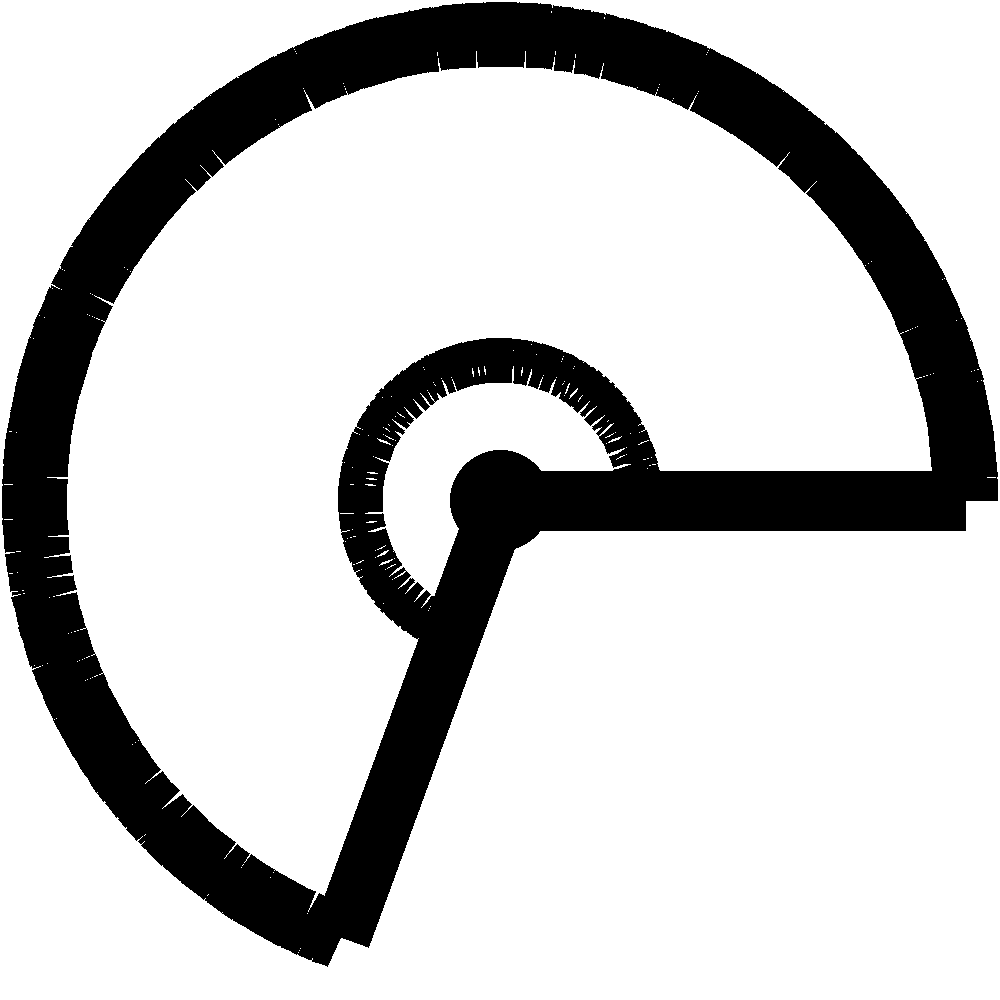}
  {Angle}
  {There is a single angle in this image.\par
   The angle is indicated by the curved arc connecting the two rays.\par
   This angle is the reference stimulus.\par
   Please assign a number to represent this angle in degrees.\par
   Use this reference for measuring other angles later.}
  {There is a single angle in this image.\par
   The angle is indicated by the curved arc connecting the two rays.\par
   Assign a number to represent this angle in degrees using the reference established earlier.}

\promptentry
  {all_figures/stimuli/area_circle_id_48.pdf}
  {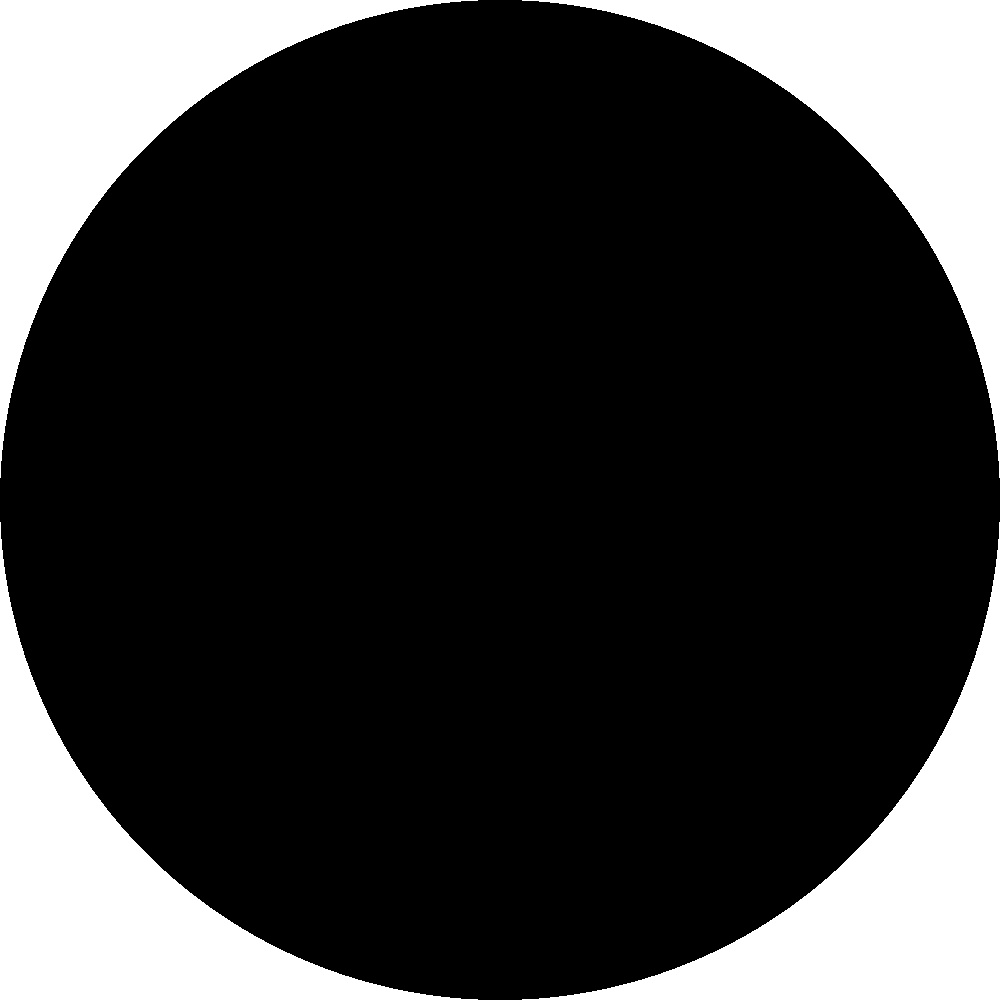}
  {Area (circular)}
  {There is a single circle in this image.\par
   This circle is the reference stimulus.\par
   Please assign a number to represent its area.\par
   Use this reference for measuring other circle areas later.}
  {There is a single circle in this image.\par
   Assign a number to represent its area using the reference established earlier.}

\promptentry
  {all_figures/stimuli/area_square_id_48.pdf}
  {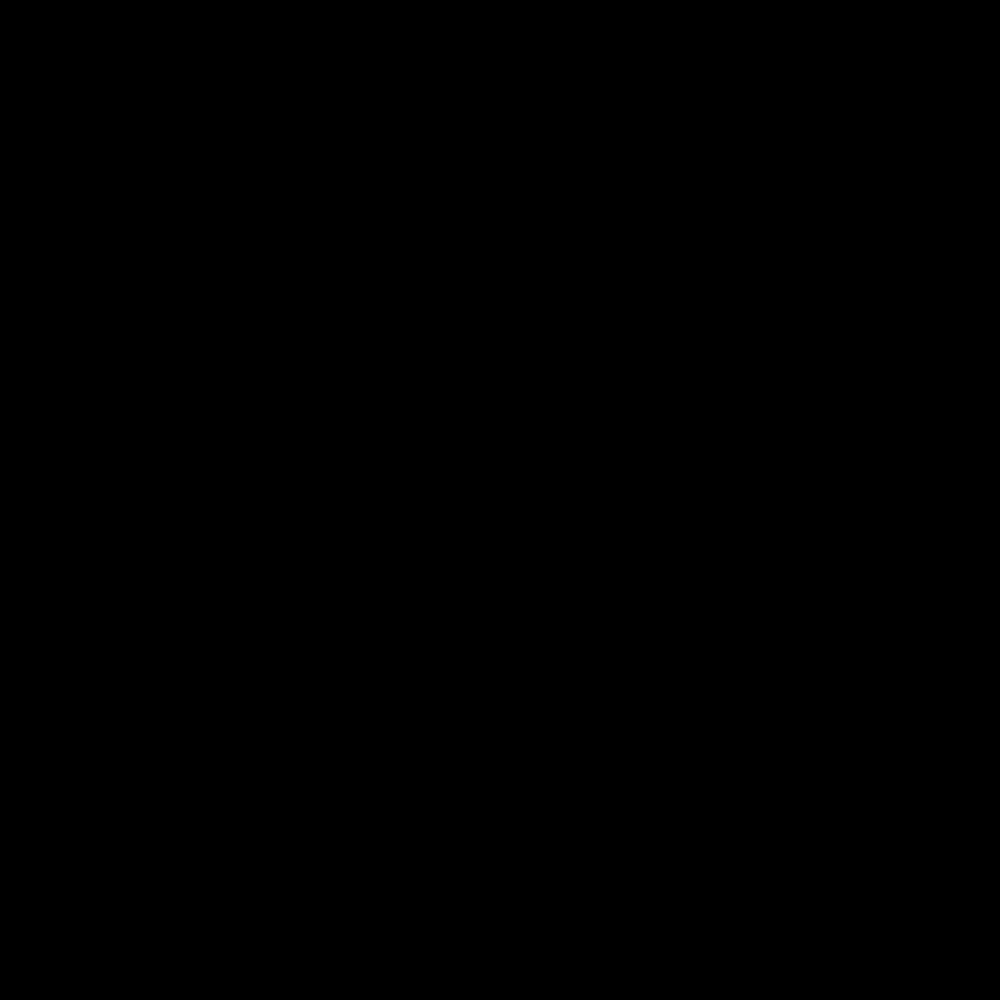}
  {Area (Square)}
  {There is a single square in this image.\par
   This square is the reference stimulus.\par
   Please assign a number to represent its area.\par
   Use this reference for measuring other square areas later.}
  {There is a single square in this image.\par
   Assign a number to represent its area using the reference established earlier.}

\promptentry
  {all_figures/stimuli/length_id_48.pdf}
  {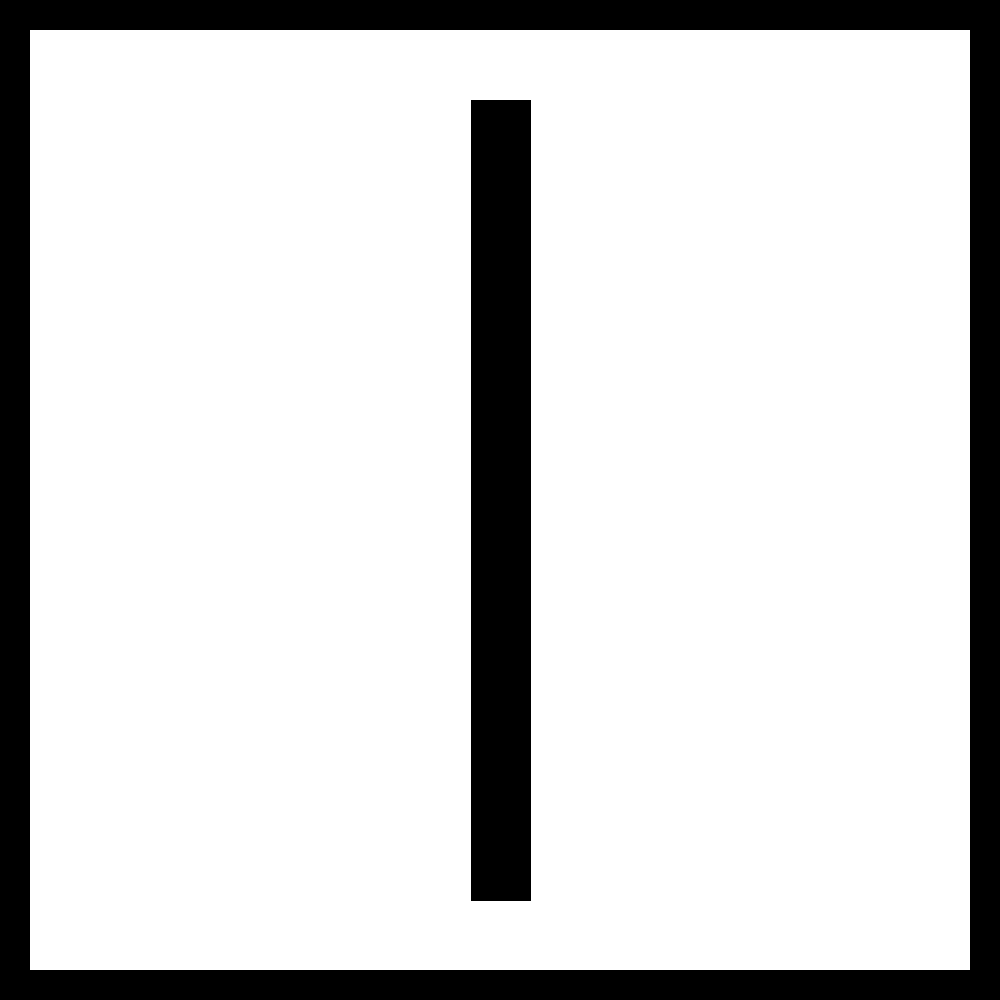}
  {Length}
  {There is a single black line in this image.\par
   This line is the reference stimulus.\par
   Please assign a number to represent its length.\par
   Use this reference for measuring other line lengths later.}
  {There is a single black line in this image.\par
   Assign a number to represent its length using the reference established earlier.}

\promptentry
  {all_figures/stimuli/slope_id_48.pdf}
  {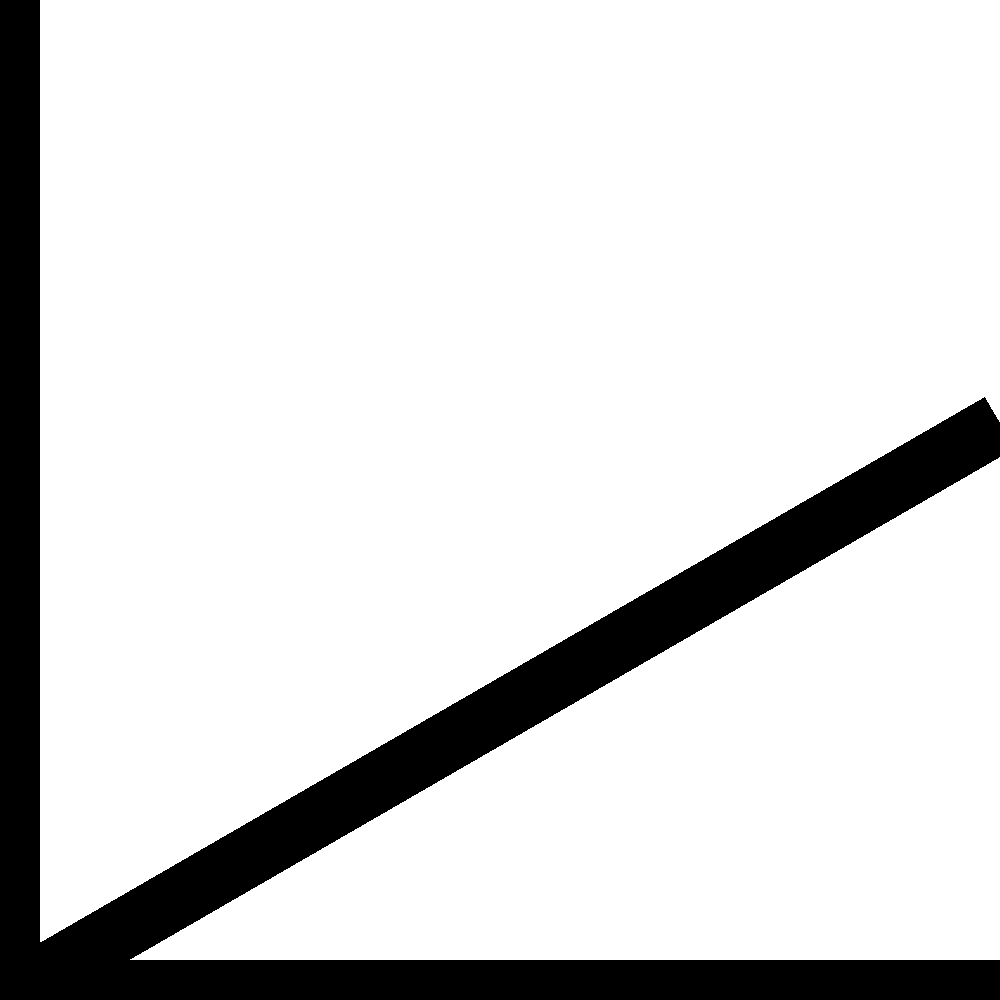}
  {Slope}
  {There is a coordinate frame with a single sloped line in this image.\par
   This sloped line is the reference stimulus.\par
   Please assign a number to represent the slope of the sloped line.\par
   Use this reference for measuring other line slopes later.}
  {There is a coordinate frame with a single sloped line in this image.\par
   Assign a number to represent this slope using the reference established earlier.}

\promptentry
  {all_figures/stimuli/shape_id_48.pdf}
  {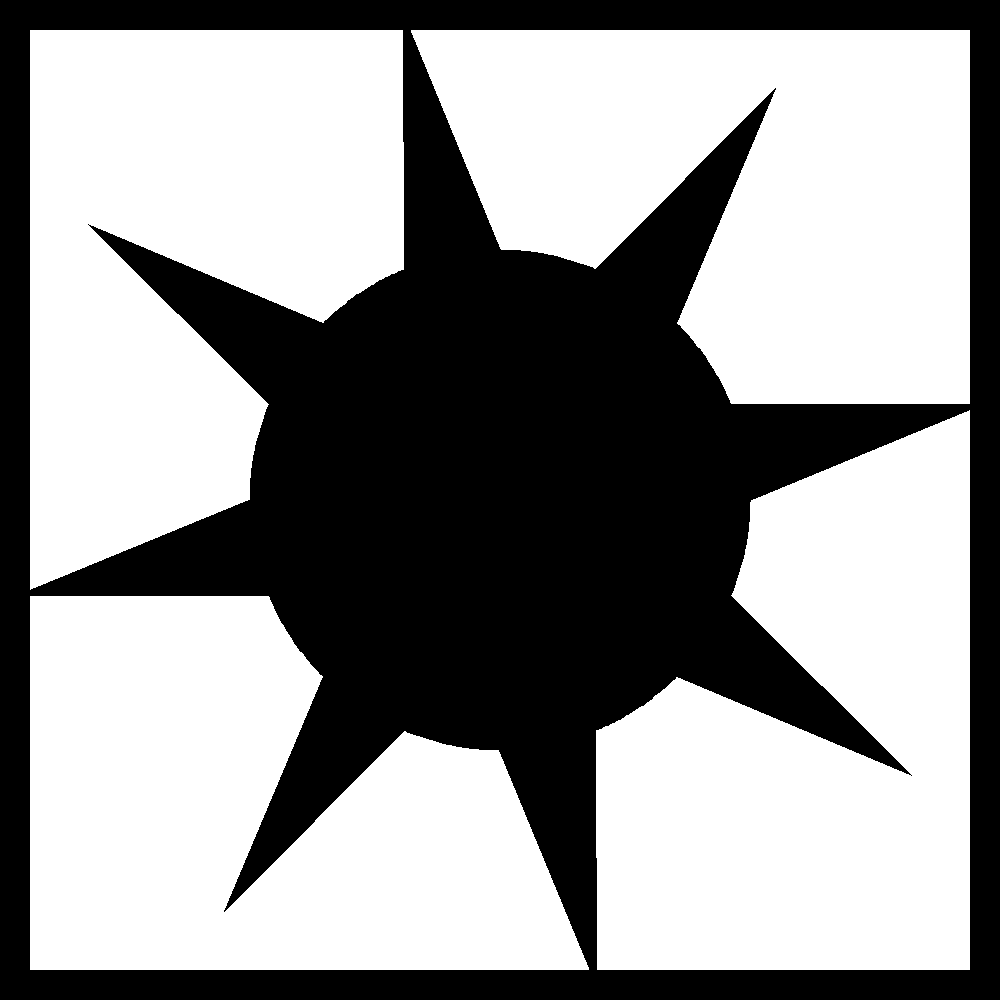}
  {Shape}
  {There is a single spiked shape in this image.\par
   This shape is the reference stimulus.\par
   Please assign a number to represent its shape.\par
   Use this reference for measuring other shapes later.}
  {There is a single spiked shape in this image.\par
   Assign a number to represent this shape using the reference established earlier.}

\promptentry
  {all_figures/stimuli/texture_dot_id_48.pdf}
  {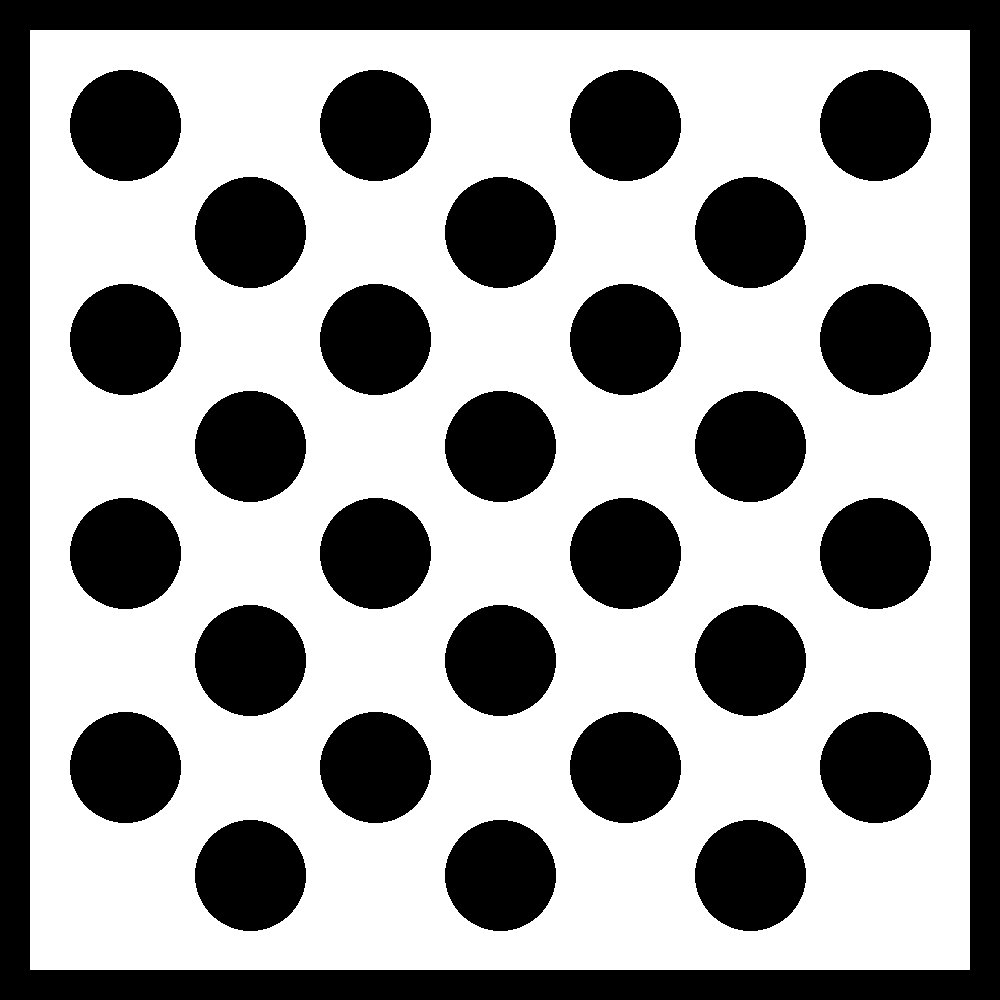}
  {Texture (point-based)}
  {There is a dotted texture in this image.\par
   This dotted texture is the reference stimulus.\par
   Please assign a number to represent this dotted texture.\par
   Use this reference for measuring other dotted textures later.}
  {There is a dotted texture in this image.\par
   Assign a number to represent this dotted texture using the reference established earlier.}

\promptentry
  {all_figures/stimuli/texture_line_id_48.pdf}
  {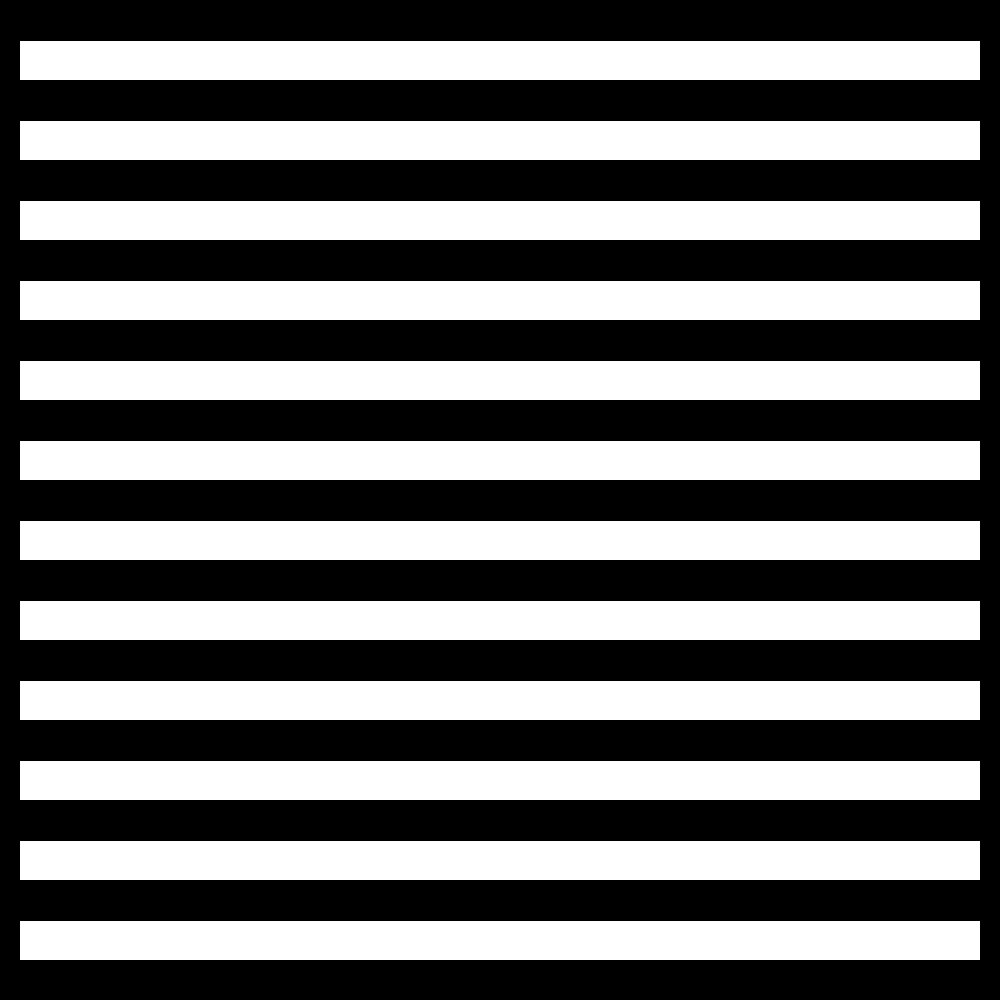}
  {Texture (line-based)}
  {There is a line texture in this image.\par
   This line texture is the reference stimulus.\par
   Please assign a number to represent this line texture.\par
   Use this reference for measuring other line textures later.}
  {There is a line texture in this image.\par
   Assign a number to represent this line texture using the reference established earlier.}

\promptentry
  {all_figures/stimuli/luminance_id_48.pdf}
  {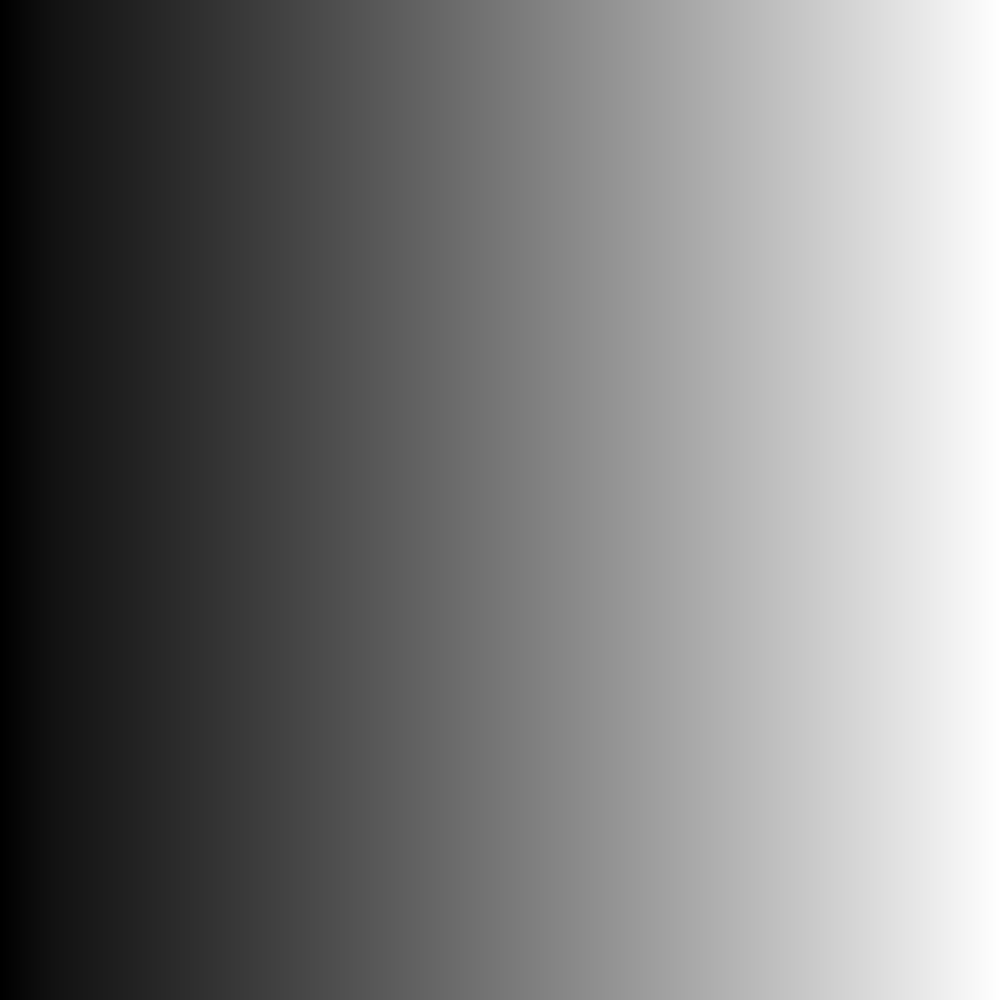}
  {Luminance}
  {There is a single colored square in this image.\par
   This colored square is the reference stimulus.\par
   Please assign a number to represent its luminance.\par
   Use this reference for measuring other luminances later.}
  {There is a single colored square in this image.\par
   Assign a number to represent its luminance using the reference established earlier.}

\promptentry
  {all_figures/stimuli/colormap_Greys_id_48.pdf}
  {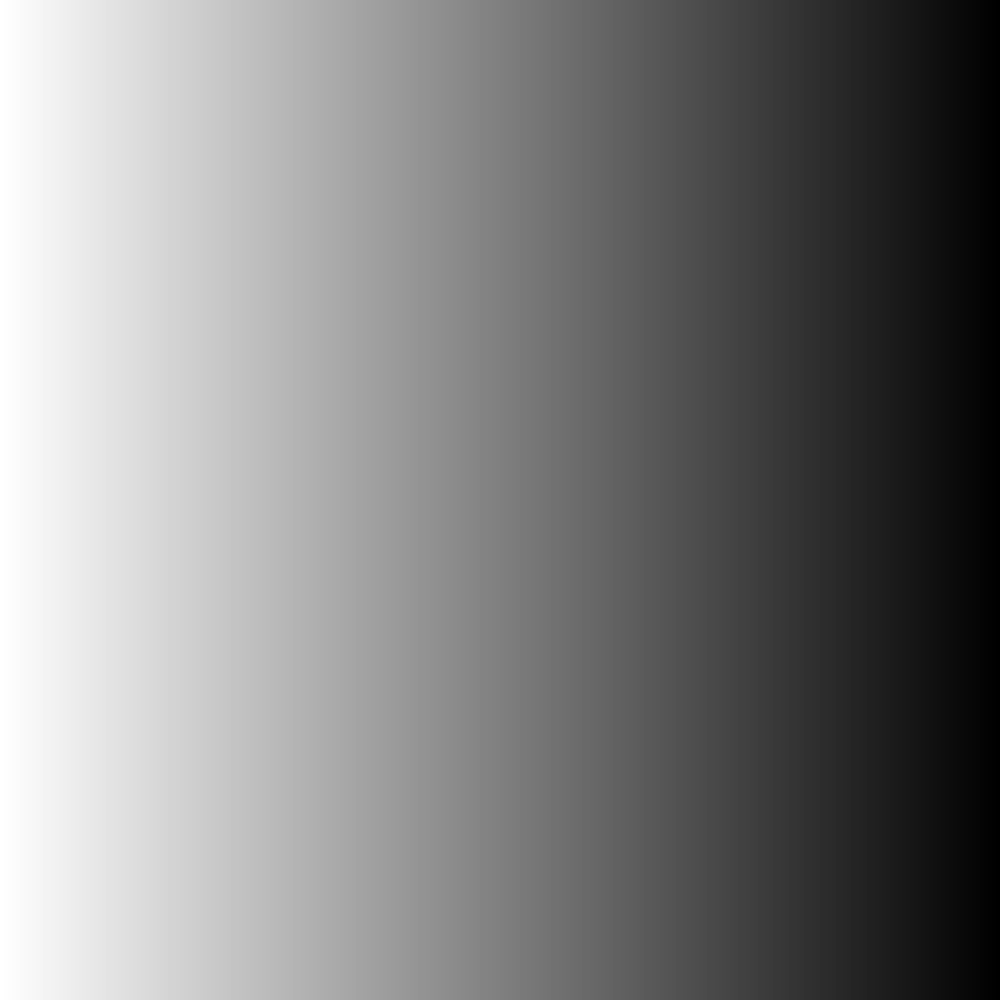}
  {Colormap (Greys)}
  {There is a single colored square in this image.\par
   This colored square is the reference stimulus.\par
   Please assign a number to represent this color.\par
   Use this reference for measuring other colors later.}
  {There is a single colored square in this image.\par
   Assign a number to represent this color using the reference established earlier.}

\promptentry
  {all_figures/stimuli/colormap_jet_id_48.pdf}
  {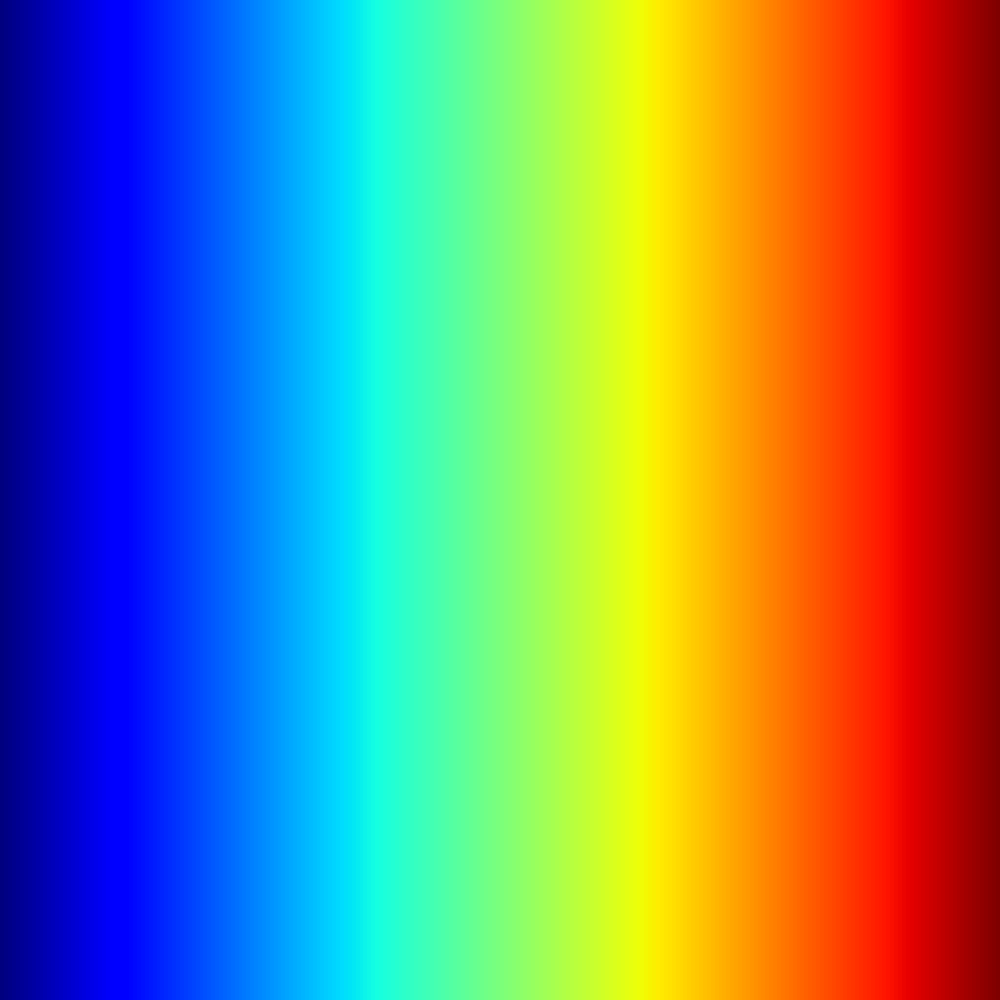}
  {Colormap (jet)}
  {There is a single colored square in this image.\par
   This colored square is the reference stimulus.\par
   Please assign a number to represent this color.\par
   Use this reference for measuring other colors later.}
  {There is a single colored square in this image.\par
   Assign a number to represent this color using the reference established earlier.}

\promptentry
  {all_figures/stimuli/colormap_plasma_id_48.pdf}
  {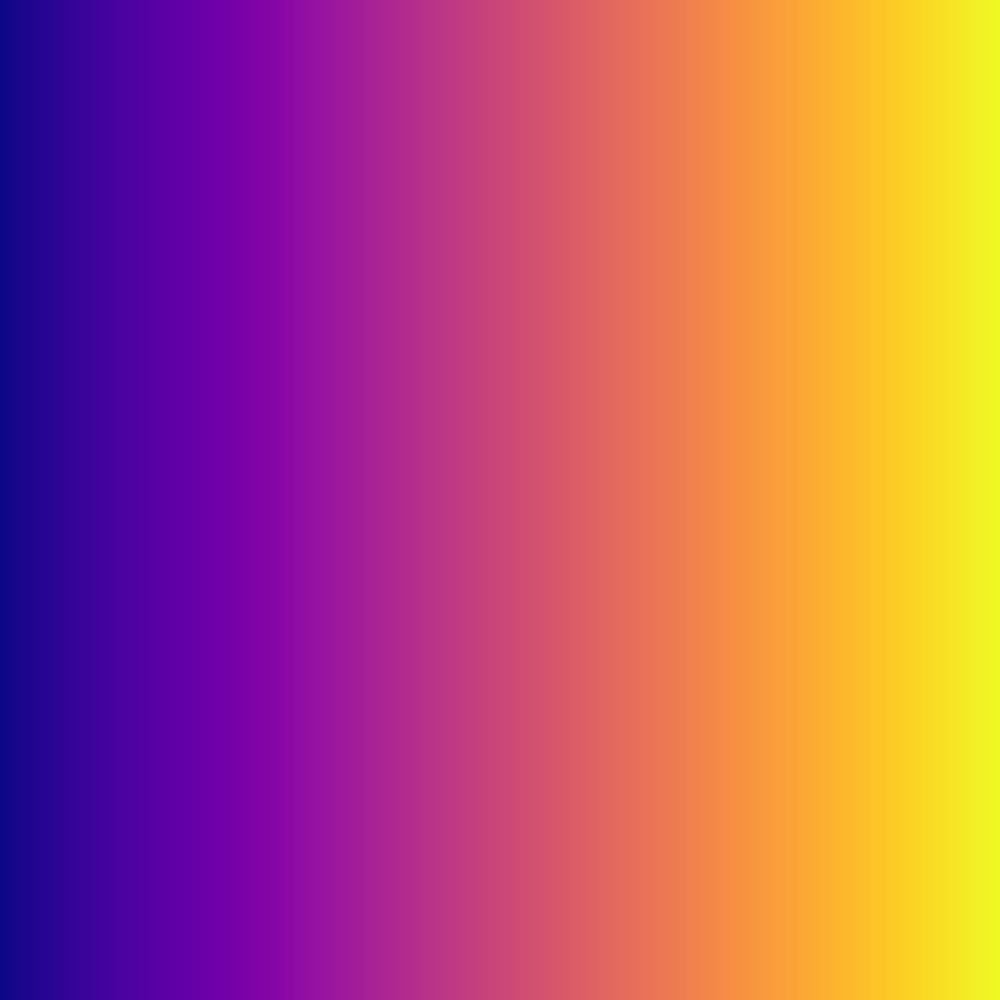}
  {Colormap (plasma)}
  {There is a single colored square in this image.\par
   This colored square is the reference stimulus.\par
   Please assign a number to represent this color.\par
   Use this reference for measuring other colors later.}
  {There is a single colored square in this image.\par
   Assign a number to represent this color using the reference established earlier.}

\section{Domain of the visual variables}
\begin{figure*}[t]
    \centering
    \includegraphics[
        width=\textwidth,
        height=0.85\textheight,
        keepaspectratio
    ]{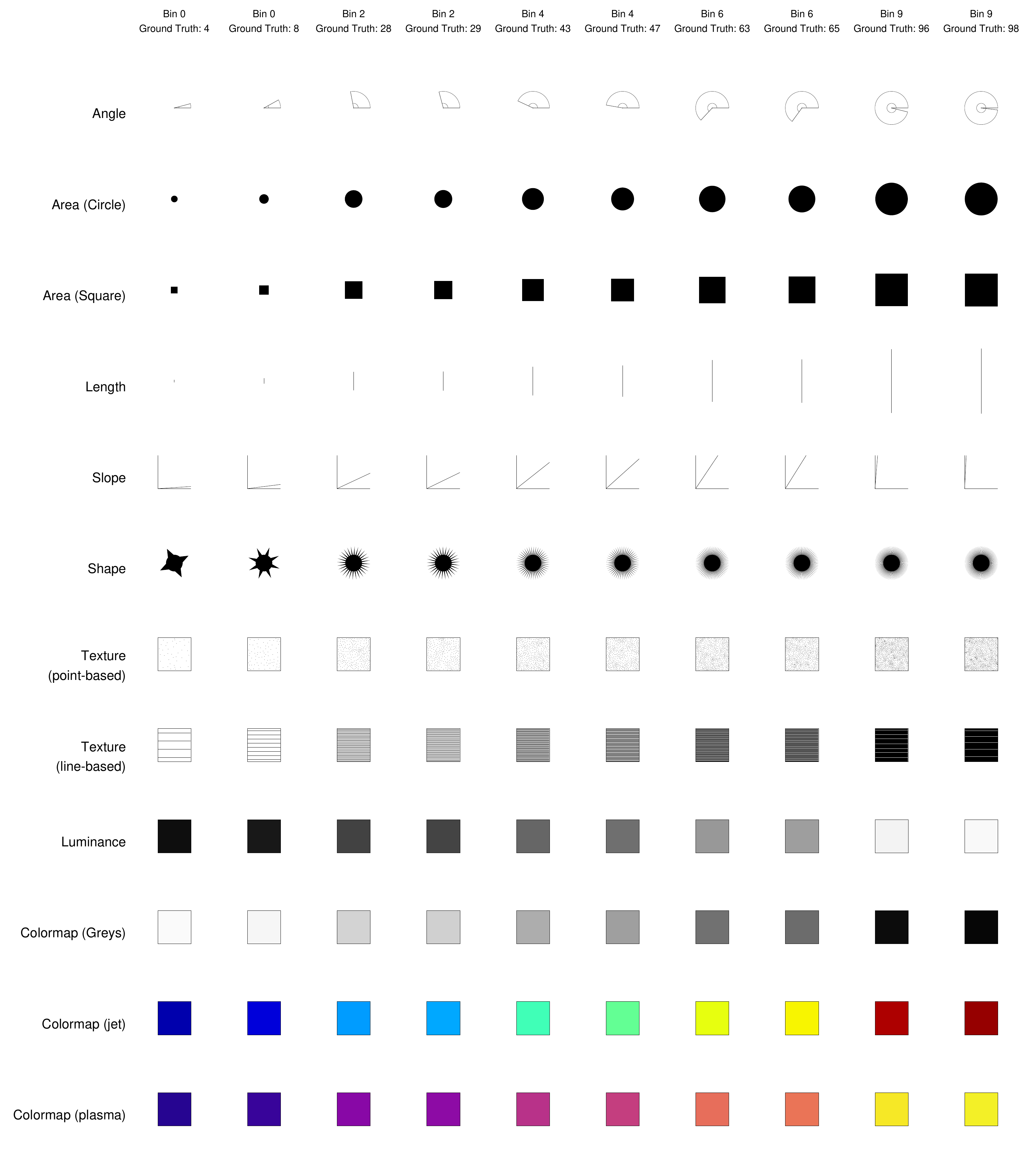}
    \caption{Ten examples for each of twelve visual variables with two examples randomly selected from each of the five sampled bins. Rows correspond to visual variables, and columns correspond to the sampled values of each bin.}
    \label{fig:stimuli-examples}
\end{figure*}

\autoref{fig:stimuli-examples} shows additional examples for the twelve visual variables.

\end{document}